\documentclass[11pt]{article}
\usepackage{tabularx}

\usepackage[preprint]{acl}
\usepackage{times}
\usepackage{latexsym}
\usepackage{graphicx}
\graphicspath{{emnlp/figures/}{figures/}}
\usepackage{float}
\usepackage{CJKutf8}
\usepackage{subcaption}
\usepackage{pifont}
\usepackage{amssymb}
\usepackage[most]{tcolorbox}
\usepackage{booktabs}

\usepackage{array}
\usepackage{amsmath}
\usepackage[utf8]{inputenc}
\usepackage[T1]{fontenc}
\usepackage{microtype}
\usepackage{xcolor}
\usepackage[normalem]{ulem}
\usepackage{seqsplit}
\usepackage{tcolorbox}
\tcbuselibrary{breakable,listings}
\lstdefinelanguage{trivialprompt}{ sensitive=false }
\tcbset{
  prompttcb/.style={
    enhanced,
    breakable,
    colback=gray!9,
    colframe=black!55,
    boxrule=0.5pt,
    arc=0.6mm,
    left=1.5mm,
    right=1.5mm,
    top=1mm,
    bottom=1mm,
    before skip=4pt,
    after skip=4pt,
    listing only,
    listing options={
      language=trivialprompt,
      basicstyle=\ttfamily\footnotesize,
      breaklines=true,
      breakatwhitespace=false,
      columns=fullflexible,
      keepspaces=true,
      aboveskip=2pt,
      belowskip=2pt,
    },
  },
}

\title{Can Mechanistic Interpretability Help Explain LLM Agent Failures?\\A Case Study on Web Element Hallucination}

\author{Mohamed Aghzal$^1$, Mor Geva$^2$, Gregory J. Stein$^1$, Ziyu Yao$^1$ \\
  $^1$George Mason University \\ 
  $^2$Tel Aviv University \\
  \texttt{\small \{maghzal, gjstein, ziyuyao\}@gmu.edu}, \small \texttt{morgeva@tauex.tau.ac.il}}

\begin{document}

\maketitle

\begin{abstract}
Mechanistic Interpretability (MI) has made significant progress in explaining the internal computations of Large Language Models (LLMs); however, these advances have yet to demonstrate comparable success in explaining the decision-making processes of LLM-based agents. This paper examines the challenges MI tools face when applied to realistic agentic failures. We focus on one such failure: web agents hallucinating actions that are not available on a website. We apply a series of mechanistic analyzes to Qwen2.5-7B-Instruct as it interacts with live websites. We find that a linear probe trained on the last prompt token can predict hallucinations before generation begins (0.80 AUC) and steering along a probe direction corrects 85\% of hallucinations, but disrupts 7\% of valid actions. However, these results only provide a partial explanation of why the agent hallucinates. The probe direction is sensitive to prompt formatting, losing much of its steering effect after a minor change to how actions are represented. Additionally, circuit identification can be difficult as the computation is distributed across multiple components and positions, and the information the agent relies on varies across examples. We discuss how these observations challenge the assumptions underlying standard MI methods for circuit discovery and feature identification, and outline future directions for making these methods better suited to explain failures in LLM-based agents.
\end{abstract}

\section{Introduction}

LLM-based agents are increasingly deployed in complex, long-horizon tasks ranging from web navigation \citep{zhou2024webarena, deng2023mind2web, xue2025an, aghzal2026llmbasedwebagentsfail} and code generation \citep{jiang2024selfplanning,tang2024worldcoder,li2025autosdt} to scientific reasoning \citep{wei2025ai,xin2024bioinformatics}. Unlike traditional LLM tasks and settings where a model produces a single response to a fixed input, agents operate sequentially: they perceive an environment, form plans, execute actions, observe outcomes, and adapt over multiple steps. While LLM-based agents have achieved success in a number of scenarios \cite{schmidgall2025agentlaboratoryusingllm, yang2024sweagent}, the complexity of underlying environments and the sequential nature of the decision-making involved introduce failure modes that are qualitatively different from those observed in standard LLM settings \citep{aghzal2026llmbasedwebagentsfail,cemri2025why,zhu2025llmagentsfaillearn,liu2024agentbench} and the mechanisms underlying these failures remain poorly understood. 

Mechanistic interpretability (MI) \citep{olah2020zoom, elhage2021mathematical} is a candidate for closing this gap. {MI aims to recover the algorithms and concepts a network implements from its weights and activations, and has traced the circuits and features responsible for a wide range of behaviors} \citep{meng2022locating, geva-etal-2021-transformer, ravfogel-etal-2021-counterfactual, finlayson-etal-2021-causal, olsson2022incontextlearninginductionheads, akyurek2023what}. While this demonstrates the promise of MI for understanding and improving LLMs, its application to LLM agents remain limited, despite agent reliability being a critical concern for both performance and safety \citep{bereska2024mechanistic, bengio2025superintelligentagentsposecatastrophic, jones2026benigninputsleadsevere, lee2026multiagentmi}.

This gap reflects a discrepancy between what current MI tools are able to explain\footnote{We use the term ``explain'' to refer not only to \emph{understanding} model internals, but also their \emph{utility}: whether the insights can be used to improve performance \citep{orgad2026interpretabilityactionable}.} and the root causes of agent failures. While MI has been designed to target model-level analysis, agentic failures emerge over long trajectories and over multiple agent components. \citet{aghzal2026chainofattribution}, thus, argue that explaining LLM agents requires multiple levels of attribution that combine behavioral analysis with MI. In this work, we focus on the mechanistic level of their proposed framework, highlighting the challenges that must be overcome to extract meaningful mechanistic insights from the backbone LLMs powering agents.

{To highlight these challenges, we conduct an analysis on a failure mode in which web agents hallucinate actions that are not available on the provided web page. Using Qwen2.5-7B-Instruct on live websites from Online-Mind2Web \citep{xue2025an}, we apply a number of MI techniques in order to explain the causes behind this behavior. We find that a linear probe on the last prompt token predicts the hallucination with 0.80 AUC before the action is written, and that steering along the probe direction fixes 85\% of hallucinations while breaking 7\% of valid actions. However, each of these analyzes captures only part of the mechanism. On the circuit discovery side, computation is distributed across components, and targeted component interventions based on component analysis fix at most 8.9\% of cases. Moreover, the input responsible varies across samples. On the feature identification side, we find that the probe direction is coupled to the prompt format and loses much of its steering effect under a small change in how actions are listed, and pretrained SAEs yield no interpretable features guiding hallucinated decisions. These findings highlight how the assumptions made by MI approaches can be challenged in the realistic settings agents operate in. We highlight some directions that can help address these issues.}

\section{Preliminaries and Background}

\subsection{The Components of an LLM Agent}\label{sec:agent-components}

LLM agents operate over interconnected components {over a sequence of timesteps $t$'s} \citep{mo2024trembling}: \textbf{Perception} converts raw observations $o_t$ into token representations $x_t = enc(o_t)$; \textbf{Memory} maintains information across timesteps through a history state (or collection of states) $h_t = u(h_{t-1}, x_t)$, taking the form of working memory (the current context window) and long-term memory (parametric knowledge and external retrieval); \textbf{Reasoning} transforms perception and memory into a decision, formalized as a policy $a_t = \pi(h_t, x_t)$, often through explicit reasoning chains \cite{wei2022chain} but can also be implicit \cite{deng2023implicit}; \textbf{Actions} $a_t \in \mathcal{A}$ are executed by an external environment through a transition function $\mathcal{E}_{t+1}=f_{\mathrm{env}}(\mathcal{E}_t, a_t)$, yielding a new observation $o_{t+1}$; and \textbf{Evaluation and Monitoring} close the loop by comparing actual and predicted outcomes, and replanning when needed.

\subsection{Mechanistic Interpretability}
{MI explains model behavior in terms of the internal mechanisms that produce it \citep{olah2020zoom, elhage2021mathematical}, studying weights, activations, and the causal relations among them instead of only inputs and outputs. Work in the area typically centers on three notions \citep{rai2024practical}. \textbf{Circuits} are subgraphs of the model that implement a recognizable algorithm, such as copying or induction, and trace how information flows to produce a given behavior \citep{ameisen2025circuittracing, meng2022locating}. \textbf{Features} are directions in activation space that encode interpretable concepts \citep{huben2024sparse, bricken2023monosemanticity}. \textbf{Universality} is the hypothesis that the same features and circuits reappear across architectures \citep{chughtai2023toymodeluniversalityreverse}.}

{Most of MI's successful explanations \citep{wang2023interpretability, hanna2023how, lindsey2025biology, geva-etal-2021-transformer, meng2022locating} come from narrow tasks with fixed inputs. Recent work has called for extending MI to more complex, realistic settings. \citet{sharkey2025open} catalogue the open problems that limit its applicability, and \citet{orgad2026interpretabilityactionable} argue that MI lacks evaluation criteria based on actionability, noting that few MI findings have so far led to concrete performance improvements. We study this question for LLM agents. Existing work is limited to sequential decision-making in toy environments or to static settings \citep{men2024unlockingfutureexploringlookahead, nanda-etal-2023-emergent, ustaomeroglu2026internal}, and the mechanisms behind agent failures in realistic environments remain largely unknown.} {A related line of work also examines the mechanisms behind LLM reasoning \cite{hou-etal-2023-towards,arcuschin2025chainofthought,dutta2024how, chen2026doescotthink}. However, agents introduce further challenges. Their failures emerge and compound over long trajectories and the computation may be spread over many components and different timesteps and positions.}

\section{The Challenges Posed by Agentic Settings for MI}\label{sec:mi_struggles}

\paragraph{Circuit Discovery}
The successful applications of circuit discovery methods \citep{wang2023interpretability, hanna2023how, conmy2023automatedcircuitdiscoverymechanistic} share implicit assumptions that can fail in agentic settings.

The first assumption is \emph{localizability}, which assumes that a small set of positions and components within a single timestep can account for the mechanisms responsible for a model's behavior. Successful applications of circuit discovery have relied on this holding consistently across samples. For instance, in subject-verb agreement \citep{finlayson-etal-2021-causal}, minimal pairs can be constructed by hand (e.g., \emph{The keys to the cabinet \textbf{are} on the table} vs.\ \emph{The key to the cabinet \textbf{is} on the table}), making it straightforward to identify the subject token and apply attribution methods that localize the relevant attention heads or FFNs as the source of the prediction.
In agentic settings, this assumption is harder to satisfy. Decisions involve complex interactions across many input positions, and there are at many times no clear basis for hypothesizing which positions or components drive an outcome and the relationship between them. As we discuss in Section~\ref{sec:web-agent-resutls}, even identifying that the action list influences hallucination in web agents leaves open the question of how different action lists give rise to different decisions. Moreover, oftentimes no single component bears a disproportionate share of the load behind the failure and many components contribute collectively in ways that single-component attribution methods can struggle to detect.

The second assumption, \emph{stationarity}, breaks down because of the temporal nature of agentic tasks: the input distribution changes at every timestep. As the agent acts and receives new observations, the context $h_t$ is continuously updated, meaning the internal representations we wish to analyze are a moving target. Standard counterfactual methods such as activation patching require a fixed clean baseline \citep{meng2022locating, wang2023interpretability}; but in an agentic setting, after each action the context window grows and the environment state changes, so no two steps share a comparable baseline and the patch no longer isolates the same computation across steps.

\paragraph{Feature Identification}

MI methods for discovering and identifying features, such as sparse autoencoders \citep[SAE;][]{bricken2023monosemanticity, huben2024sparse}, probing classifiers \citep{belinkov-2022-probing, hewitt-manning-2019-structural, tenney2019bert}, and steering vectors \citep{zou2023representation, turner2024steeringlanguagemodelsactivation, postmus2024steering, rimsky-etal-2024-steering}, face challenges similar as circuit discovery.

The first challenge concerns the \emph{data distribution.} SAEs are trained to decompose activations into interpretable features. However, the features they recover reflect what is prevalent in their training corpus. Thus, without agent-specific data, SAEs may struggle to capture concepts relevant to agentic behavior such as goal tracking, tool selection, or multi-step planning, which arise from environment interaction rather than text. Prior work already shows that SAE performance degrades under covariate shift \citep{kantamneni2025are, pacela2026stopprobingstartcoding}, which naturally indicates that limited robustness is to be expected when applying them to agent trajectories. 
Probing classifiers \citep{belinkov-2022-probing, tenney2019bert} face a related issue: the linear structure they rely on can vary dramatically over the course of a multi-turn conversation \citep{lampinen2026linearrepresentationslanguagemodels}, and can be brittle in the face of distribution shifts \citep{pacela2026stopprobingstartcoding}. As such, the nature of agent trajectories, which involve dynamic context accumulation across steps and can vary greatly depending on design choices as well as environments, means that probing-based approaches can be unreliable.
Similarly, steering vectors assume the identified direction transfers to the target setting, but distribution shifts can move the representations out of the subspace the intervention vector was designed on, making it ineffective.

The second difficulty concerns the \emph{temporal nature} of agent failures. Agent decisions are not necessarily encoded in a single feature at a single timestep. Commitment to a particular plan or strategy may be distributed across multiple features and emerge gradually over the course of a trajectory \citep{men2024unlockingfutureexploringlookahead,dong2025emergent}, with no single timestep at which the decision is cleanly represented. \citet{lubana2026priors} illustrate this limitation, showing that standard SAEs struggle to encode temporal features even in textual settings. In stateful, long-horizon environments, this issue can be more severe, as decisions and their representations can depend on trajectory-level structure.

\section{Why do Web Agents Hallucinate Website Elements?}

Hallucination \citep{zhang2025sirenssongaiocean} is an issue that has been widely documented in LLMs, where systems generate plausible but factually incorrect information. This issue presents critical safety concerns for agents \citep{deng2025threats, zhang2024toolbehonest,zhu2025llmagentsfaillearn,lin2025llmbasedagentssufferhallucinations, yona2026hallucinationsunderminetrustmetacognition}. In the context of web agents, this could arise as proposing website actions that reference nonexistent website components \citep{aghzal2026llmbasedwebagentsfail}. In this case study, we aim to explain this behavior. {While the case study is specific to web agent hallucination, the challenges we illustrate can also apply to other agents, as we discuss in Appendix \ref{app:other-agents}}.

\begin{figure*}[t]
\centering
\resizebox{\linewidth}{!}{$
\overbrace{\texttt{TYPE}}^{\texttt{action}}\quad
\overbrace{
  \texttt{//input[contains(@placeholder,\,"}\,%
  \underbrace{
\texttt{Search make, model, or keyword}
  }_{\texttt{selector value}}
  \texttt{"\,)]}
}^{\texttt{selector}}
\quad\overbrace{\texttt{Tesla Model 3}}^{\texttt{input value}}
$}
\vspace{-0.1in}
\caption{Example hallucination. Given the task: \emph{\textbf{Find a 2022 Tesla Model 3 on CarMax.}}, the model generates a \texttt{TYPE} action targeting a search input. The selector value does not correspond to a valid action provided on the webpage. A valid action would be, e.g., 
\texttt{CLICK  //a[contains(@id, 'header-drawer-focus-start')](text:"Shop")}, which clicks the "Shop" link in the header to open inventory browsing.}
\label{fig:selector_example}
\end{figure*}

\subsection{Experimental Setup} 

When restricted to a minimal action space of \texttt{CLICK}, \texttt{SCROLL}, \texttt{TYPE}, and \texttt{HOVER} (following the original Mind2Web \citep{deng2023mind2web}, an early benchmark of web navigation tasks), even frontier models have been shown to fabricate actions referencing nonexistent page elements \citep{aghzal2026llmbasedwebagentsfail}. We leverage Online-Mind2Web \citep{xue2025an}, which collects interaction episodes on live websites. On each webpage interaction (a maximum of 10 per task), we construct a list of possible actions based on the web page and provide this action list to the model to choose from, using the prompts in Appendix \ref{app:prompts}. We then define a \emph{hallucinated action} as one where the agent generates an action that does not appear in the list; an example is shown in Figure \ref{fig:selector_example}. 

We then conduct an analysis on the \texttt{Qwen2.5-7B-Instruct} \citep{qwen2025qwen25technicalreport}. 
 We observe that such hallucination is often due to the model committing to a strategy that does not align with the website; the most common pattern consists of actions targeting elements such as \texttt{Search}, \texttt{Explore}, or \texttt{Find}, that do not exist on the page but reflect the model's intent, as shown by logit lens analysis (Appendix~\ref{app:logit_lens}). The final dataset consists of 300 episodes and 2,071 actions, 32.6\% of which are hallucinated.

\subsection{The Challenges of Circuit Discovery}\label{sec:web-agent-resutls}

We first demonstrate the challenges of attributing hallucinations to a set of positions and components.

\begin{table}[t]
\centering
\small
\setlength{\tabcolsep}{3pt}
\caption{Intervention results using different sets of attention heads and FFN layers ($\alpha=3$, $\beta=0.5$).}
\vspace{-0.1in}
\label{tab:intervention}
\resizebox{\columnwidth}{!}{
\begin{tabular}{@{}lccc@{}}
\toprule
\textbf{Attn Heads} & \textbf{FFN Layers} & \textbf{Fix} (\textuparrow) & \textbf{Break} (\textdownarrow)  \\
\midrule
L22 (All heads) & 22, 23, 24 & \textbf{8.9\%} & \textbf{2.4\%} \\
L22, L27 (All heads) & 22, 23, 24 & 5.9\% & 27.6\% \\
Top-10 causal attr. (Figure \ref{fig:attn}) & 13, 15, 24 & 4.4\% & 3.5\% \\
Top-10 causal attr. (Figure \ref{fig:attn}) & 22, 23, 24 & 4.5\% & 4.6\% \\
Top-10 corr.\ diff. (Figure \ref{fig:attn-difference})& 22, 23, 24 & 6.3\% & 2.5\% \\
\bottomrule
\end{tabular}
}
\end{table}

\begin{figure*}[t]
    \centering
    
    \begin{subfigure}{\linewidth}
        \centering
        \includegraphics[width=\linewidth]{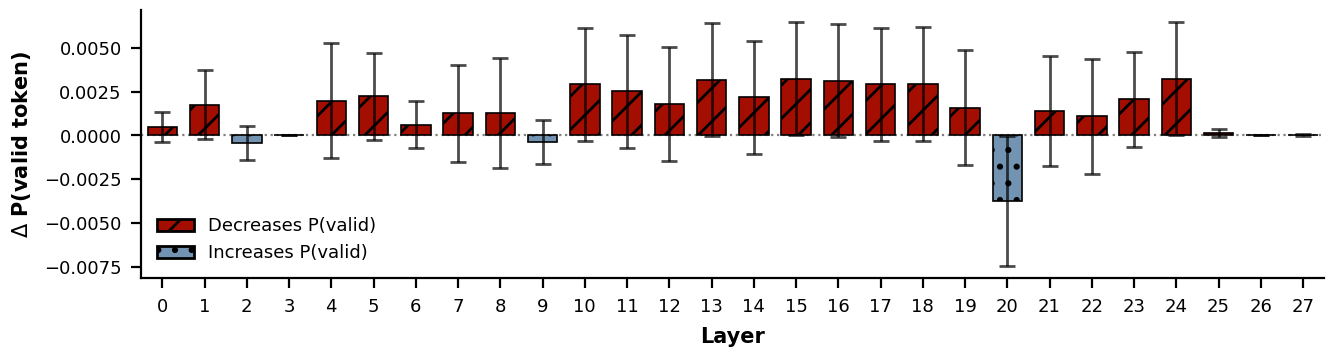}
        \caption{
        \textbf{Per-layer FFN amplification effect on valid actions.} For each FFN layer independently, we amplify its output by $10\times$ before addition to the residual stream and measure $\Delta$P(valid token) $=$ P(valid $\mid$ clean) $-$ P(valid $\mid$ FFN$_l$ amplified). Red bars: amplification reduces P(valid), i.e., the FFN pushes the model away from valid page tokens. Blue bars: amplification increases P(valid), i.e., the FFN contributes to grounding.}
        \label{fig:ffn}
    \end{subfigure}
    \hfill
    \begin{subfigure}[t]{0.49\linewidth}
        \centering
        \includegraphics[width=\linewidth]{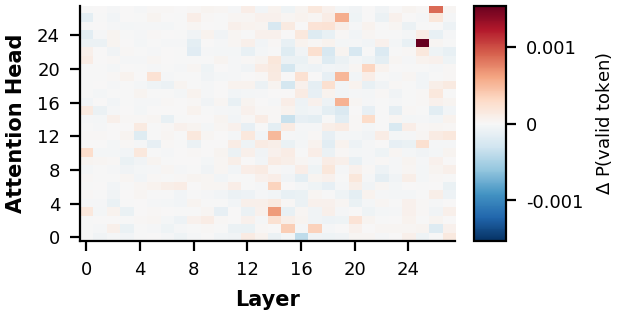}
        \caption{\textbf{Per-head attention noise effect on grounded actions.} Gaussian noise is injected into each head's projection independently. $\Delta$P(valid) $=$ P(valid$\mid$clean) $-$ P(valid$\mid$noised head), averaged over grounded cases at selector value start. Red: noising reduces P(valid); blue: noising increases P(valid).}
        \label{fig:attn}
    \end{subfigure}
    \hfill
    \begin{subfigure}[t]{0.49\linewidth}
        \centering
        \includegraphics[width=\linewidth]{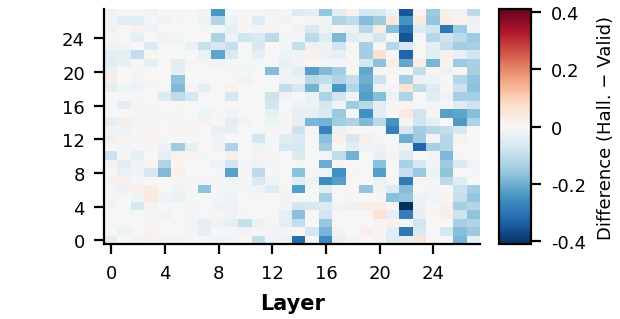}
        \caption{
        \textbf{Per-head max attention value difference (hallucinated $-$ valid) at the selector value start.} For each head, we compute the maximum attention weight placed on any token, averaged over actions per-class, and plot the difference. Blue $=$ higher in valid actions; red $=$ higher in hallucinated actions. }
        \label{fig:attn-difference}
    \end{subfigure}

    \caption{Component contributions to hallucination. (a) Per-layer FFN amplification effect. (b) Per-head attention noise effect. (c) Per-head max value difference across all heads. Setup details in Appendix~\ref{app:single-comp}.}
    \label{fig:component-analysis}
   
\end{figure*}

\paragraph{\uline{Localizability (of responsible LM components) breaks down}: Web element generation is a largely distributed computation.} Prior work has shown that hallucinations in LLMs can often be attributed to FFN contributions overshadowing attention head outputs in the residual stream, and that targeted dampening of specific FFNs while amplifying attention heads can reduce them \citep{sun2025redeep}. We test whether this holds in our setting by amplifying FFN contributions 10-fold one at a time and applying Gaussian noise to individual attention heads to valid cases {at \texttt{selector\_value\_start}} and assessing whether the outputs become hallucinated {and show the results in Figure~\ref{fig:component-analysis}}. The results show that most FFNs have little effect or contribute negatively to grounding, and most attention heads contribute minimally to valid token probability. However, while interventions on these individual components do produce hallucinations, they are qualitatively different from the hallucinations we get in zero-shot settings, which are more descriptive of strategies. For example, as shown in the examples in Appendix~\ref{app:intervention_failures}, amplifying the FFN at layer 15 transforms a valid selector value \texttt{interface jsx-326752785 wiki-section bold underlined} 
into {\ttfamily\seqsplit{VAR-1000000000...}}, a syntactically invalid sequence that does not clearly attempt to achieve a non-existent strategy. This is in contrast to several natural hallucinations such as \texttt{Search make, model, or keyword}, which are syntactically well-formed but reference non-existent elements. This suggests that no individual component contributes significantly to such decisions.

We further test joint interventions similarly to \citet{sun2025redeep}, scaling sets of FFNs by a coefficient \(\beta\) and attention heads by a coefficient \(\alpha\). As shown in Table~\ref{tab:intervention}, targeted interventions on the top\nobreakdash-10 attention heads identified and the 3 FFNs that consistently lead to hallucinations yield only modest gains (4.4\% fix rate, 3.5\% break rate). The best configuration among an extensive set of runs (see Appendix~\ref{app:intervention_full} for details), amplifying all attention heads at layer 22 while halving FFN contributions at layers 22--24, fixes 8.9\% of hallucinated cases but corrupts 2.4\% of valid ones, despite none of these components (with the exception of L24) leading significantly to hallucinations when intervened on individually. {Correlational analysis (Figure~\ref{fig:component-analysis}c)} also confirms that layer 22 attention heads are the most differentially activated between grounded and hallucinated cases. Nevertheless, although this confirms that these components have a role in grounded generation, they do not tell the full story; if the computation were localized to these components, targeted interventions should produce consistent improvements \citep{sun2025redeep,meng2022locating,wang2023interpretability}. The fact that they do not, and that simultaneously high fix and break rates occur across configurations, suggests that the hallucinations are distributed and coupled in a way that single-component attribution cannot explain.

\paragraph{\uline{Localizability (of responsible input positions) breaks down.} The action list is a contributor, but the specific positions and properties responsible for hallucination remain unidentifiable.} {As detailed in Appendix~\ref{app:causal_trace_search_full}, activation patching shows the action list as a causal contributor (Table~\ref{tab:causal-tracing-search}), with clean state restoration at L27 recovering up to 0.36 probability mass after corruption compared to less than 0.02 for \texttt{selector\_start} and the task description. However, unlike tasks such as factual recall where replacing \textit{Eiffel Tower} with \textit{Colosseum} produces a clean shift from \textit{Paris} to \textit{Rome} \citep{meng2022locating}, knowing the action list matters does not reveal which of its properties drives the commitment, leaving no clear handle for intervention. In fact, even the input span to corrupt is not stable across samples. {Corrupting the action list of valid steps with Gaussian noise, as in corruption-based methods such as activation patching and EAP-IG \citep{hanna2024have},} changes the model's decision only when the action list is its sole source, whereas in the half of valid steps where the agent had already used the same value in an earlier step, it recovers it from its action history, as can be seen in Figure~\ref{fig:corruption-valid}.

\begin{table}[t]
\centering
\setlength{\tabcolsep}{3pt}
\caption{Activation patching results for $P(\texttt{Search})$. Columns are corruption spans; ``Corrupted $P$'' is the mean $P(\texttt{Search})$ before restoration. Rows show the peak causal recovery $\Delta P(\texttt{Search})$ per component, with peak layer and token offset.}
\vspace{-0.1in}
\label{tab:causal-tracing-search}
\resizebox{\columnwidth}{!}{%
\begin{tabular}{lccc}
\toprule
 & \textbf{Selector start} & \textbf{Task description} & \textbf{Action list} \\
\midrule
\textbf{Corrupted $P$} & 0.682 & 0.672 & \uline{0.325} \\
\midrule
\textbf{Residual}      & 0.013\ (L6,\ $-$4)  & 0.012\ (L27,\ \ 0) & \textbf{0.360}\ (L27,\ \ 0) \\
\textbf{Attention}     & 0.009\ (L27,\ 0)    & 0.011\ (L23,\ \ 0) & \textbf{0.075}\ (L22,\ \ 0) \\
\textbf{FFN}           & 0.008\ (L18,\ 0)    & 0.009\ (L18,\ \ 0) & \textbf{0.067}\ (L21,\ \ 0) \\
\bottomrule
\end{tabular}%
}
\vspace{-0.2in}
\end{table}

\begin{figure}[t]
\centering
\includegraphics[width=\columnwidth]{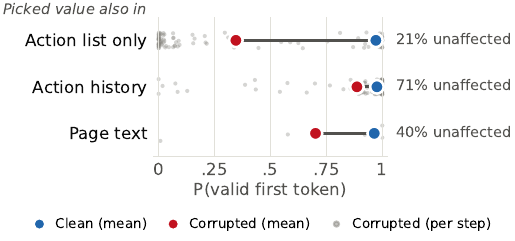}
\caption{Effect of corrupting the action list on valid steps, grouped by where else the selector value the agent picked appears in the prompt. Dots show the mean probability $P$ of a valid first token at \texttt{selector\_value\_start} before (blue) and after (red) corruption. \% of actions that remain unaffected after each corruption is shown.}
\label{fig:corruption-valid}
\vspace{-0.1in}
\end{figure}

\paragraph{\underline{Stationarity is violated}: Attempting to construct counterfactuals introduces cascading divergence instead of a single controlled difference.} 
{A natural extension would be to apply activation patching directly to paired valid and hallucinated trials; however, this requires structurally matched prompts differing in one controlled dimension. In our setting this is not possible: Different actions come from entirely different pages, tasks, URLs, and action histories and the two instances can vary dramatically. As shown in Appendix~\ref{app:counterfactual_divergence}, constructing pairs by removing the chosen valid action yields a hallucinated output in only 38\% of cases, and produces cascading trajectory divergence, rendering the two runs incomparable after a single step.}

\subsection{The Challenges of Feature Identification}

Next, we illustrate the challenges of feature identification in agentic settings.

\paragraph{\underline{Temporal locality is violated}: The hallucination decision can already be encoded pre-generation.} 
We test whether the distinction between hallucinated and valid actions corresponds to a stable internal representation. We train linear probes on the residual stream at three input positions: the end of the prompt (\texttt{prompt\_end}), i.e. before the model starts generation, the start of the selector (\texttt{selector\_start}), and the start of the selector value (\texttt{selector\_value\_start}). Probes are trained independently at each of the 28 layers using an episode-level split (70/15/15 of 300 episodes, yielding 1,433 training actions, 282 development, and 356 test). As shown in Figure~\ref{fig:probe}, all three positions yield probes that perform remarkably well on the test set. Interestingly, at \texttt{prompt\_end}: a probe trained on pre-generation activations reaches a test AUC of 0.80 at layer 18, before the model has produced a single output token. By the time we get to \texttt{selector\_start} the signal becomes much stronger even at layer 0 and by \texttt{selector\_value\_start} the signal is strong from the first layer and only mildly improves throughout. This suggests that the model's commitment to a hallucinated action is already encoded as a linearly separable direction in the residual stream before generation begins, and may be influenced by context accumulated across previous timesteps and several tokens before the hallucination emerges. This violates temporal locality: rather than being encoded at the timestep where the hallucination manifests, the commitment is already present in the residual stream before generation begins and strengthens progressively across positions. 

\paragraph{\underline{Distributional mismatch of linear probing}: The direction identified can be coupled to the agentic design.} To further test whether the hallucination-predictive direction in the residual stream is causally upstream of the output, we apply representation steering using the weights of the trained probe (See Appendix \ref{app:steering} for more details and detailed results). This intervention helps substantially; reducing hallucinations by as much as 85\%, while only shifting 7\% of valid cases to hallucinations using the probe from L17 at \texttt{selector\_start}.\footnote{While this confirms a causal role at this position, it does not satisfy the localizability assumption: which specific components or input properties drive the effect remains unclear.}
However, when modifying the format of the action list slightly, these probes are brittle to distributional shift: replacing the JSON action list with a numbered markdown format (Appendix~\ref{app:ood-prompt}) causes probe performance to degrade substantially (Figure~\ref{fig:ood-probes}) and the steering effectiveness to drop significantly despite the trained probe having seen the data otherwise as shown in Figure~\ref{fig:steering-ood-bar}. This suggests that the learned direction is entangled with action format instead of encoding only the hallucination direction.

\begin{figure}[t]
    \centering
    \includegraphics[width=0.89\columnwidth]{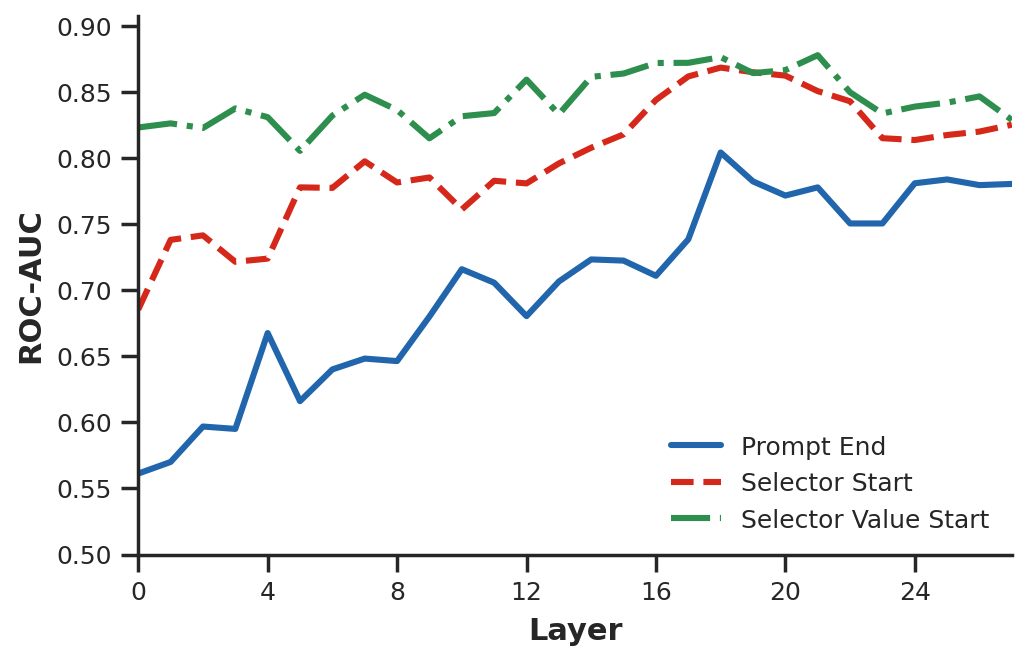}
    \caption{Per-layer AUC of linear probes trained to predict hallucination at different positions.}
    \label{fig:probe}
    \vspace{-0.1in}
\end{figure}

\begin{figure}[t]
    \centering
    \includegraphics[width=0.85\linewidth]{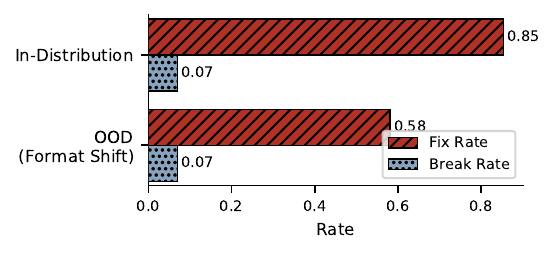}
        \vspace{-0.15in}
    \caption{Steering results for the \texttt{selector\_start} L17 vector.}
    \label{fig:steering-ood-bar}
    \vspace{-0.1in}
\end{figure}

\begin{figure*}[t]
    \centering
    \includegraphics[width=\linewidth]{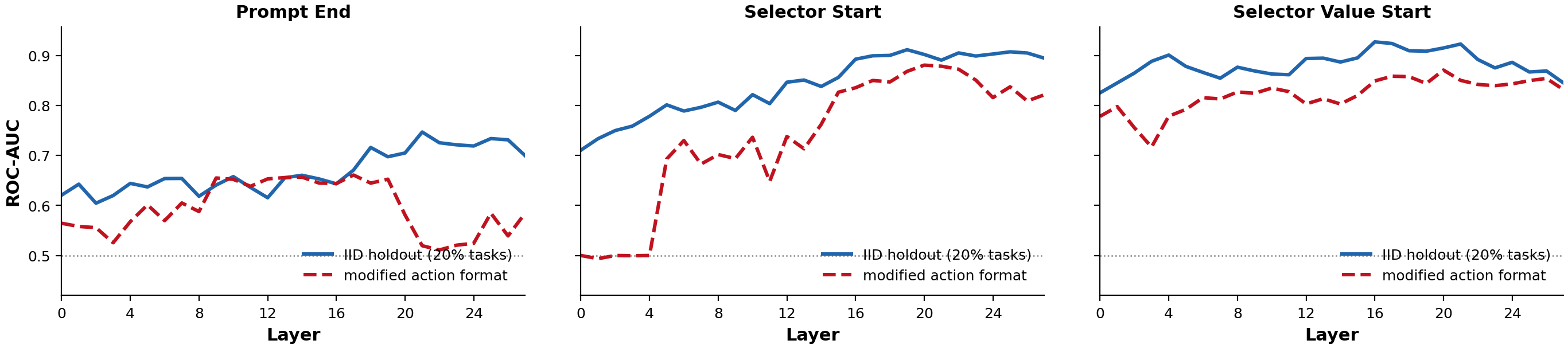}
    \caption{Per-layer ROC-AUC of probes trained to predict hallucination after modifying the action format.}
    \label{fig:ood-probes}
\end{figure*}

\paragraph{\uline{Distributional mismatch of SAEs}: Pretrained SAEs do not yield clearly interpretable features in a web agent setting.} 

{We also apply a pretrained SAE to to the action list span at layer 27; in order to indentify if any features relevant to hallucination can be extracted.  However, the top features differentiating hallucinated from valid cases mostly promote tokens unrelated to the observed hallucination behavior (Figure~\ref{fig:sae}), highlighting how SAEs pretrained on text can struggle to produce interpretable directions for features that emerge only during agent interaction.}

\begin{figure}[t]
    \centering
    \includegraphics[width=\linewidth]{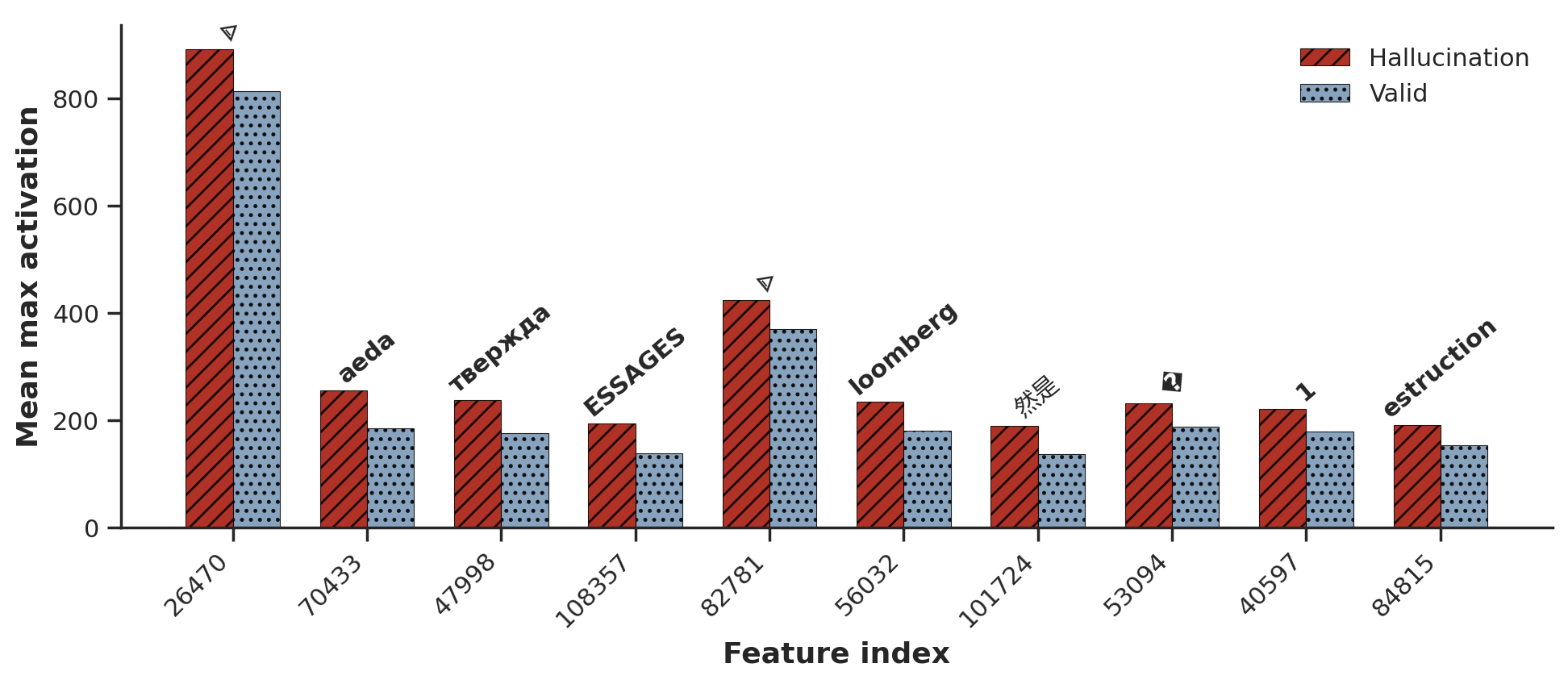}
    \caption{Top 10 SAE features by absolute difference in how strongly they activate in hallucinated vs valid cases at L27 and the token most promoted by each direction.}
    \label{fig:sae}
\end{figure}

\section{A Roadmap for MI for LLM Agents}\label{sec:roadmap}

We believe that MI methods to explain LLM agents need methods to address the following.

\paragraph{{MI Benchmarks for Agentic Applications}}
MI Benchmarks  \citep{wang2023interpretability, mueller2025mib, karvonen2025saebench, gao2025scaling} have played an important role in grounding progress in concrete, reproducible tasks; however, no {MI} benchmarks exist for {agent applications}. A benchmark for agentic MI must specify not just a task and a model but an environment, a trajectory distribution, and a ground truth for both a trajectory-level failure (as opposed to a surface-level one) and what triggered such failures. In our case study, for example, nothing indicated which input the agent relied on for a given decision. It must also define what a correct mechanistic explanation looks like and provide a way to evaluate whether a proposed explanation is faithful, complete, and actionable. We believe that developing such benchmarks is {crucial} to measure progress and compare approaches.

\paragraph{Move Beyond Single {Position/}Component Causal Attribution}
Many MI methods \citep{conmy2023automatedcircuitdiscoverymechanistic, syed-etal-2024-attribution, ameisen2025circuittracing, dai2022knowledgeneuronspretrainedtransformers} assume failures can be localized to a small number of components that bear disproportionate causal responsibility. However, in agentic settings, computation is distributed across many components and positions, making such causal methods inapplicable. {In our case study, the best joint intervention on attention heads and FFNs fixes 8.9\% of hallucinations (Table~\ref{tab:intervention}), and that no intervention on a single component can reproduce the hallucination pattern (Figure \ref{fig:component-analysis}).} {Recent work has started to tackle} attribution over multi-token responses \citep{pan2026longhorizoninterpretabilityefficientfaithful} and identify intervention points for distributed behaviors \citep{sankaranarayanan2026activation}, however, methods capable of attributing failures distributed across multiple timesteps and environment states remain scarce. 

\paragraph{Construct Counterfactuals in Stateful Environments}
Building clean interpretable counterfactuals in realistic agentic settings is rarely possible: each trajectory is shaped by a unique sequence of environment states, agent decisions, and observations, meaning that no two trials are structurally comparable in the way that activation patching requires. As our case study illustrates, valid and hallucinated actions can come from entirely different pages, tasks, and URLs and can occur at different timesteps. {Removing the chosen valid action yields a hallucination in only 38\% of cases and diverges the trajectory within one step (Appendix~\ref{app:counterfactual_divergence}), and even within a single step, corrupting the action list leaves the agent's choice unchanged when the selected element also appears in its history as shown in Figure~\ref{fig:corruption-valid}.} Addressing this requires new methods for constructing or approximating counterfactuals in dynamic environments. While some literature does exist for this problem \cite{tsirtsis2021counterfactual, triantafyllou2025counterfactual}, {they assume known state spaces and comparable trajectories, which is not the case in open-ended environments with unbounded action spaces and highly variable observations.} Thus, methods that are applicable to realistic agent environments are needed.

\paragraph{Feature Discovery for Agentic Distributions}
{SAEs \citep{lieberum-etal-2024-gemma, templeton2024scaling, bloom2024opensource} are typically trained on text corpora, therefore, their features reflect text rather than interaction, and they may miss concepts that arise only when a model acts in an environment.} {This is illustrated by the fact that the top SAE features separating hallucinated from valid cases in our case study promote tokens unrelated to the behavior (Figure~\ref{fig:sae}), and the probe direction that does predict hallucination fails to transfer when only the action format changes (Figure~\ref{fig:ood-probes}).} {Concepts such as tracking a goal, committing to a multi-step strategy, and modeling how the environment behaves are unlikely to be captured in textual corpora alone. Additionally, they do not have clear counterparts in current feature antologies \citep{neuronpedia2023}, and they may not be well captured by the linear directions SAEs search for \cite{bhalla2026sparseautoencoderscaptureconcept}. Progress will, thus, require SAEs trained on agent trajectories, methods that capture richer geometric structure, and feature ontologies tailored toward agents, which would give researchers shared terms for finding and evaluating the features involved in agent decisions.}

\paragraph{Monitoring Model Beliefs Across Timesteps}

{Even when a failure is clearly observable at the output, it is not straightforward to determine when the model committed to the wrong strategy or how its understanding of the environment evolved to that point.} Fully addressing this requires methods that can track the emergence of a plan across both layers and timesteps and identify the point at which (or if) the model's internal state transitions from open exploration to a fixed plan and which (if any) implicit heuristics the model adopts to select candidate decisions. 
{This is harder than in static settings because the commitment may crystallize gradually across multiple timesteps, and the context that triggers it is shaped by a history of environment interactions that differs across every trajectory.} A line of recent work \cite{yalon2026indicationsbeliefguidedagencymetacognitive, lubana2026priors, lepori-etal-2025-racing, lampinen2026linearrepresentationslanguagemodels} examines belief evolution in LLMs, but agentic settings introduce unique challenges such as long-horizon dependencies, tool interactions, and action-observation feedback loops. 

\paragraph{Establishing Universality Across Agent Types}
A foundational hypothesis in MI is that many features and circuits recur across models trained on similar tasks \citep{wang2025towards, chughtai2023toymodeluniversalityreverse}, suggesting that mechanistic insights can generalize beyond the specific model and setting on which they were discovered. Whether a similar form of universality holds across agent types remains an open question. For instance, if a {circuit responsible for planning} can be identified in a web agent, does an equivalent one appear in an embodied or a code generation agent? If such {findings} are universal across agent types, then the field can accumulate knowledge across settings. 
\section{Conclusion}

In this paper, we investigate how far standard MI tools can go in explaining agent failures in realistic settings. Focusing on web element hallucination, we find that these tools can detect an agent’s decision before it acts and steer it away from failure, but cannot explain how that decision is made. Each tool falls short for a reason tied to properties of agents that challenge the assumptions underlying existing MI methods. We therefore argue that making MI useful for understanding agents will require methods specifically designed to account for these properties.

\section*{Acknowledgment}
This project was sponsored by the National Science Foundation (NSF) Grant No. 2418580 and 2540626 and the Foresight Institute. This material is also based upon work supported by NSF under Grant No. 2232733. G. J. Stein acknowledges support from this grant. G. J. Stein is also supported by Army Research Lab under award W911NF2520011. We also thank the George Mason NLP Group for their helpful comments. This project was also supported by resources provided by the Office of Research Computing at George Mason University (URL: https://orc.gmu.edu) and funded in part by grants from the National Science Foundation (Award Number 2018631).

\newpage

\bibliography{bibliography}


\appendix

\newpage

\section{How Our Case Study Results Generalize Beyond Web Agents}\label{app:other-agents}

Although the results of our case study are specific to web navigation, the same MI challenges arise across agent settings. In embodied settings, agents struggle with path and motion planning even in simple environments \citep{aghzal2024can, aghzal2024look, chen2024autotamp}, and their decisions can be driven by unreliable representations that produce plausible but environmentally ungrounded reasoning chains \citep{aghzal2024evaluating, ma2024agentboard, liu2024agentbench}, leading to degenerative cycles in which an agent commits to an action sequence that violates environmental preconditions and corrupts subsequent belief states \citep{yuan2026rpmsenhancingllmbasedembodied, zhu2025llmagentsfaillearn}. In tool calling agents, hallucinations arise not from generation errors but from incorrect tool selection strategies that later steps cannot recover from \citep{peng2026tool}. In code generation agents, failures are distinguished not by individual generation errors but by early diagnostic mistakes that propagate into repair strategies \citep{liu2025empiricalstudyfailuresautomated, majgaonkar2025code}. Across all three settings, the causal origin of failure can be distributed across the trajectory, the environment state changes at every timestep, and failure is jointly determined by model computation and external dynamics, which are the same difficulties our case study identifies.

\section{Prompt Template for Data Synthesis}
\label{app:prompts}

Agent trajectories analyzed in this work are produced by rolling out the model on Online-Mind2Web tasks. The following prompts are used.

\paragraph{System message.}
\begin{tcblisting}{prompttcb}
You are a web navigation agent. Always respond with exactly two lines: [THOUGHT] and [ACTION]. The [ACTION] must be a valid JSON object from the available actions list.
\end{tcblisting}

\paragraph{User message (step $t=0$).}
Variables \texttt{<TASK>}, \texttt{<DOM>}, and \texttt{<ACTIONS\_JSON>} denote the dataset task string, the DOM summary, and the formatted action list, respectively.
\begin{tcblisting}{prompttcb}
Task: <TASK>

You are given a task to perform on a webpage. Propose the next step that is helpful towards achieving the task.

You should output only an action that is most likely to help achieve the subgoal. Each action is a dictionary with the following keys:

{
  "ACTION": "CLICK" | "HOVER" | "TYPE" | "SELECT",
  "SELECTOR": "text='NEWS'" | "button:nth-of-type(3)" | "css=div.menu >> text='Super Bowl'" | ...,
  "VALUE": "if any (You may need to provide a value for the action, e.g. for TYPE action)",
  "TEXT": "the text that is visible on the element",
  "EXPLANATION": "a short explanation of why this action is useful"
}

Below is the simplified representation of the current state of the webpage:
<DOM>

Below is the list of available actions. DO NOT OUTPUT ANY ACTIONS THAT ARE NOT IN THE LIST OF AVAILABLE ACTIONS. THE ONLY FIELD YOU ARE ALLOWED TO MODIFY IS THE "VALUE" (if a TYPE or SELECT action) AND "EXPLANATION" FIELDS. YOU ARE NOT ALLOWED TO MODIFY ANY OTHER FIELDS.
IT IS VERY IMPORTANT THAT YOUR OUTPUT CONTAINS ONLY THE ACTION, AND NOTHING ELSE, THAT IS DIRECTLY PARSABLE AS A JSON OBJECT.

Available Actions:
<ACTIONS_JSON>
=========END OF ACTIONS=========

Continue with:
[THOUGHT] <one short sentence>
[ACTION] <JSON action object from the list of available actions>
\end{tcblisting}

\paragraph{User message (step $t>0$).}
The prompt repeats the task, then up to the last three stored steps as lines \texttt{Step $k$: [THOUGHT] ... [ACTION] ... Verdict: ...}, then the current DOM and the same available-actions constraint and \texttt{Available Actions:} block as above.

\section{Logit Lens Analysis}
\label{app:logit_lens}

We use the logit lens \citep{nostalgebraist2020logitlens} to decode the model's intermediate layer representations into vocabulary predictions at the selector value position. As shown in Figure~\ref{fig:logit-lens}, by layers 21--27, hallucinating trials converge on a stable set of semantically coherent action labels (\emph{Search}, \emph{Enter}, \emph{Find}, \emph{Explore}) that are plausible website UI elements but do not exist on the current page. Interestingly, the model commonly produces the Chinese word for ``Search'' (\begin{CJK}{UTF8}{gbsn}搜索\end{CJK}) at L24 before converging on the English word, indicating that the model commits to strategies not grounded in the specific website. Valid trials at these late layers converge on the correct tokens for the page. This divergence suggests that in hallucinated cases, the model has resolved a generic concept of what the action should look like to achieve the desired intent rather than grounding its prediction in the specific page.

\begin{figure*}[t]
    \centering
    \includegraphics[width=\linewidth]{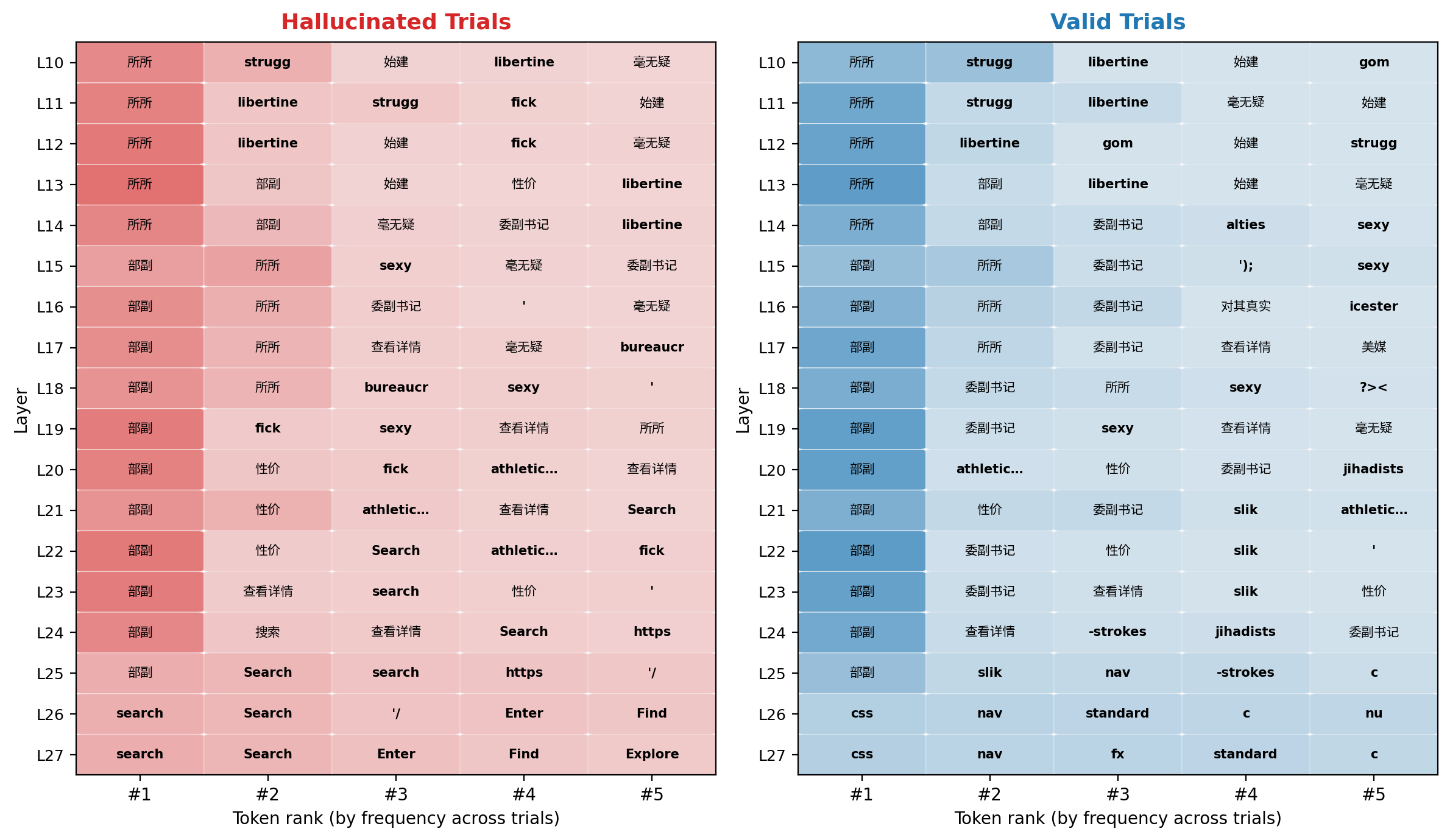}
    \caption{Top-5 most frequent tokens after logit lens projection for mid-to-late layers across valid and hallucinated cases. Cell intensities indicate how often each word appears as the top token at the corresponding layer.}
    \label{fig:logit-lens}
\end{figure*}

\section{Single Component Attribution Results}\label{app:single-comp}

We evaluate two forms of single-component attribution at the \texttt{selector\_value\_start} position. For FFN attribution, we amplify each layer's FFN output by $10\times$ before adding it to the residual stream, and measure $\Delta$P(valid token) $=$ P(valid~$\mid$~clean) $-$ P(valid~$\mid$~FFN$_l$ amplified) averaged over grounded actions. For attention head attribution, we inject independent Gaussian noise into each head's output projection and measure the same quantity. We additionally compute a correlational signal by comparing, for each head, the maximum attention weight placed on any token across valid vs.\ hallucinated actions.

As shown in Figure~\ref{fig:ffn}, most FFN layers have negligible or mild effects on $P(valid)$, with the notable exception of layer 20 which contributes positively to grounding. Attention head effects (Figure~\ref{fig:attn} are similarly diffuse: most heads cause only small changes to $P(valid)$, showing that no single component bears a disproportionate share of the causal load. However, the correlational difference heatmap in \ref{fig:attn-difference} shows that layer 22 attention heads are the most differentially activated between classes, meaning that these heads contribute jointly to grounding; which is in line with observed results in Tables \ref{tab:intervention} and \ref{tab:intervention_full}.

\section{Intervention Failure Examples (Before vs. After)}
\label{app:intervention_failures}

We show representative \emph{valid\(\rightarrow\)hallucinated} examples. Each example starts from a selector value that is valid and becomes invalid after intervention.

\begin{sloppypar}
\paragraph{Configuration A: \(L25H23\times 4\), \(L20\times 2\).}
\textbf{Aggregate effect:} \(3.6\%\) valid trials broken.

    \textbf{Before (baseline, valid):} \texttt{filterpanel-accordion-button-0}\\
    \textbf{After (intervened, hallucinated):} \texttt{filterpanel-8-gender-1} \\
    \textbf{Before (baseline, valid):} \texttt{job-seekers}\\
    \textbf{After (intervened, hallucinated):} \texttt{for-job-seekers}

\paragraph{Configuration B: \(L25H23\times 5\), \(L20\times 5\).}
\textbf{Aggregate effect:} \(74.9\%\) valid trials broken.

    \textbf{Before (baseline, valid):} \texttt{group/cta flex min-h-8 w-fit flex-row items-center gap-1 pr-4 text-neutral} \\
    \textbf{After (intervened, hallucinated):} \texttt{group-cto flex min-h-3.5 min-h-3.65 h-3.5 h-6 flex-1 flex-1.5 flex-1.5} \\
    \textbf{Before (baseline, valid):} \texttt{sc-cSUYbb hFerSt}\\
    \textbf{After (intervened, hallucinated):} \texttt{sc-cI336o gZdZdZ}

\paragraph{Configuration C: \(L25H23\times 10\), \(L20\times 10\).}
\textbf{Aggregate effect:} \(99.9\%\) valid trials broken.

    \textbf{Before (baseline, valid):} \texttt{Link\_link\_\_1AZfr SubNav\_subNav\_\_link\_\_54qyo}\\
    \textbf{After (intervened, hallucinated):} {\ttfamily\seqsplit{SubsideItem\_\_item\_\_1000000000000000000000000000000000}} \\
    \textbf{Before (baseline, valid):} \texttt{silver}\\
    \textbf{After (intervened, hallucinated):} {\ttfamily\seqsplit{oneoff Silver +100000000000000000000000000000000000}}

\paragraph{Direct FFN amplification.}

\textbf{Layer 15}:

    \textbf{Before (valid):} \texttt{silver}\\
    \textbf{After (amplified):} \texttt{\_silver}\\
    \textbf{Before (valid):} \texttt{interface jsx-326752785 small}\\
    \textbf{After (amplified):} \texttt{guides}\\
    \textbf{Before (valid):} \texttt{interface jsx-326752785 wiki-section bold underlined}\\
    \textbf{After (amplified):} {\ttfamily\seqsplit{VAR-1000000000000000000000000000000000000}}

\textbf{Layer 26}:

    \textbf{Before (valid):} \texttt{link text-p-400 no-underline hover:underline}\\
    \textbf{After (amplified):} \texttt{discussion}\\
    \textbf{Before (valid):} \texttt{interface jsx-326752785 small}\\
    \textbf{After (amplified):} \texttt{slick-slide slick-active}\\
    \textbf{Before (valid):} \texttt{interface jsx-326752785 wiki-section bold underlined}\\
    \textbf{After (amplified):} \texttt{slick-slide slick-current slick-active}
\end{sloppypar}

\section{Additional Intervention Results}
\label{app:intervention_full}

Table~\ref{tab:intervention_full} reports additional intervention configurations sorted by net improvement (Fix\%~$-$~Break\%). 

\begin{table*}[t]
\centering
\caption{Head-selective intervention runs sorted by Net $\Delta$\% (Fix \% - Break \%). $\alpha$: attention head scale. ``—'' = no intervention on the specific component (\texttt{selector\_value\_start)}.}
\label{tab:intervention_full}
\begingroup
\setlength{\tabcolsep}{3.5pt}
\scriptsize
\centering
\begin{tabular}{p{4.55cm}p{4.05cm}rrrr}
\toprule
\textbf{Attn Heads} & \textbf{FFN} & $\boldsymbol{\alpha}$ & \textbf{Fix\%} & \textbf{Brk\%} & \textbf{Net\%} \\
\midrule
    L22~(all) & L22--L24$\times$0.5 & 3 & 8.90 & 2.37 & 6.53 \\
    L22~(all) & L22--L23$\times$0.5 & 3 & 7.95 & 2.03 & 5.92 \\
    L22~(all) & L0,L5,L22--L24$\times$0.5 & 3 & 8.33 & 2.54 & 5.80 \\
    L22~(all) & L22--L23$\times$0.5, L20$\times$1.5 & 3 & 7.01 & 1.52 & 5.48 \\
    L22~(all) & L22$\times$0.5, L20$\times$1.5 & 3 & 5.87 & 1.02 & 4.86 \\
    L22~(all) & L22$\times$0.5 & 3 & 6.25 & 2.37 & 3.88 \\
    L22~(all) & L22$\times$0.8 & 3 & 5.49 & 1.69 & 3.80 \\
    L22~(all) & — & 3 & 5.11 & 1.35 & 3.76 \\
    L21H5, L22H1,3,4,6,7,23,24,27, L26H14 & L22--L24$\times$0.5 & 3 & 6.25 & 2.54 & 3.71 \\
    L20H21, L21H5, L22H4,6,7,23,24,27, L26H14 & L13,L15,L24$\times$0.5 & 3 & 4.73 & 1.86 & 2.87 \\
    L22~(all) & L22--L24,L27$\times$0.5 & 3 & 7.77 & 5.75 & 2.01 \\
    L25H23 & L23--L24$\times$0.5 & 2 & 2.08 & 0.17 & 1.91 \\
    L22~(all) & L22--L27$\times$0.5 & 3 & 8.33 & 7.11 & 1.23 \\
    L14H3,6,12, L15H1, L17H1, L19H16,19,26, L25H23, L26H27 & L13,L15,L24$\times$0.5 & 3 & 4.36 & 3.55 & 0.80 \\
    L22H0,1,3,4,24,27 & — & 5 & 5.30 & 4.57 & 0.73 \\
    L22H4, L25H23 & — & 3 & 1.70 & 1.02 & 0.69 \\
    L25H23 & L10--L19,L23--L24$\times$0.8 & 3 & 1.89 & 1.35 & 0.54 \\
\midrule
    L14H3,6,12, L15H1, L17H1, L19H16,19,26, L25H23, L26H27 & L22--L24$\times$0.5 & 3 & 4.55 & 4.57 & $-$0.02 \\
    L22~(all) & L22$\times$0.8 & 5 & 7.58 & 7.61 & $-$0.04 \\
    L22~(all) & — & 5 & 7.39 & 7.45 & $-$0.06 \\
    L14H3,6,12, L15H1, L17H1, L19H16,19,26, L25H23, L26H27 & L13,L15,L24$\times$0.5, L20$\times$1.5 & 3 & 3.98 & 4.57 & $-$0.59 \\
    L14H3,6,12, L15H1, L17H1, L19H16,19,26, L25H23, L26H27 & — & 3 & 3.60 & 4.23 & $-$0.63 \\
    L22H0,4,24,27, L25H23 & — & 5 & 3.60 & 4.23 & $-$0.63 \\
    L22H4, L25H23 & L10,L15,L24$\times$0.8, L20$\times$1.2 & 5 & 2.46 & 3.21 & $-$0.75 \\
    L14H3,6,12, L15H1, L17H1, L19H16,19,26, L25H23, L26H27 & L23--L25$\times$0.5, L20$\times$1.5 & 3 & 4.17 & 5.08 & $-$0.91 \\
    L22H0,4,24,27, L25H23 & L15$\times$0.8 & 5 & 3.03 & 4.40 & $-$1.37 \\
    L22H0,1,3,4,24,27 & — & 5 & 5.30 & 6.77 & $-$1.47 \\
    L25H23 & L20$\times$2 & 4 & 1.33 & 3.21 & $-$1.89 \\
    L22H0,1,3,4,24,27, L25H23 & L20$\times$2 & 5 & 5.11 & 7.45 & $-$2.33 \\
    L22H1,4,22,24,27, L25H23 & L15$\times$0.8 & 5 & 4.92 & 7.61 & $-$2.69 \\
    L25H23 & L10--L19,L23--L24$\times$0.5 & 1.5 & 2.65 & 6.60 & $-$3.95 \\
    L25H23, L26H26 & L20$\times$3 & 3 & 2.84 & 9.31 & $-$6.47 \\
    L25H23 & L20$\times$3 & 3 & 2.65 & 9.64 & $-$6.99 \\
    L22H0,1,3,4,24,27, L25H23 & L20$\times$3 & 3 & 3.60 & 11.51 & $-$7.91 \\
    L25H23 & L20$\times$3 & 10 & 2.84 & 22.67 & $-$19.83 \\
    L22H1,3,4,24,27 & L5$\times$0.5 & 15 & 7.33 & 27.75 & $-$20.42 \\
    L22~(all), L27~(all) & — & 3 & 5.87 & 27.58 & $-$21.71 \\
    L22H0,1,3,4,24,27, L25H23 & L20$\times$3 & 10 & 5.49 & 38.58 & $-$33.09 \\
    L22H0,1,3,4,24,27, L25H23 & — & 15 & 4.55 & 62.61 & $-$58.06 \\
    L22H0,1,3,4,24,27, L25H23 & L20$\times$3 & 15 & 5.30 & 69.04 & $-$63.73 \\
    L25H23 & L20$\times$5 & 5 & 1.14 & 73.60 & $-$72.47 \\
    L25H23 & L20$\times$10 & 10 & 0.19 & 100.00 & $-$99.81 \\
    — & L15$\times$10 & 1 & 0.00 & 100.00 & $-$100.00 \\
    — & L10--L19,L23--L24$\times$10 & 1 & 0.00 & 100.00 & $-$100.00 \\
    \bottomrule
\end{tabular}
\endgroup
\end{table*}


\section{Activation Patching Results for \texttt{Search}}
\label{app:causal_trace_search_full}

To identify which parts of the input encode the commitment to a hallucinated action, we apply activation patching \citep{meng2022locating} and focus on the specific cases in which the model generated \texttt{Search} at the start of the selector value. For each such case, we corrupt a candidate span with Gaussian noise, then restore the model's clean hidden states at each layer and token offset pair and measure the recovery in the probability of the search token $P(\texttt{Search})$ at the selector value. We report the results in Table~\ref{tab:causal-tracing-search}, and how the full causal-trace heatmaps (all traced layers and offsets) in Figures \ref{fig:search-trace-1}, \ref{fig:search-trace-2} and \ref{fig:search-trace-3}. Each panel reports the mean intervention effect on \(p(\texttt{Search})\) for residual, attention, and FFN restoration. All results are reported across three spans: the selector start, the task description, and the action list.

While all three corruptions reduce $P(\texttt{Search})$ to some degree, the action list accounts for the majority of recoverable causal signal: corrupting it produces the largest drop in $P(\texttt{Search})$ and restoring clean states at L27 recovers up to 0.36 probability mass. Notably, while the residual stream at L27 yields the strongest single restoration effect, attention and FFN individually contribute far less (0.075 and 0.067, respectively), suggesting that the information is distributed across these components rather than in either mechanism alone. However, identifying the action list as a contributor is only a partial result: unlike linguistic tasks such as factual recall, where knowing the subject token is causal immediately suggests a targeted intervention \citep{meng2022locating} (e.g. replacing the subject \textit{Eiffel Tower} with \textit{Colosseum} produces a shift from \textit{Paris} to \textit{Rome}), knowing that the action list matters does not explain which of its properties drives the commitment to the decision, making the result less interpretable and the intervention less straightforward.

\begin{figure*}[t]
    \centering
    \includegraphics[width=0.32\linewidth]{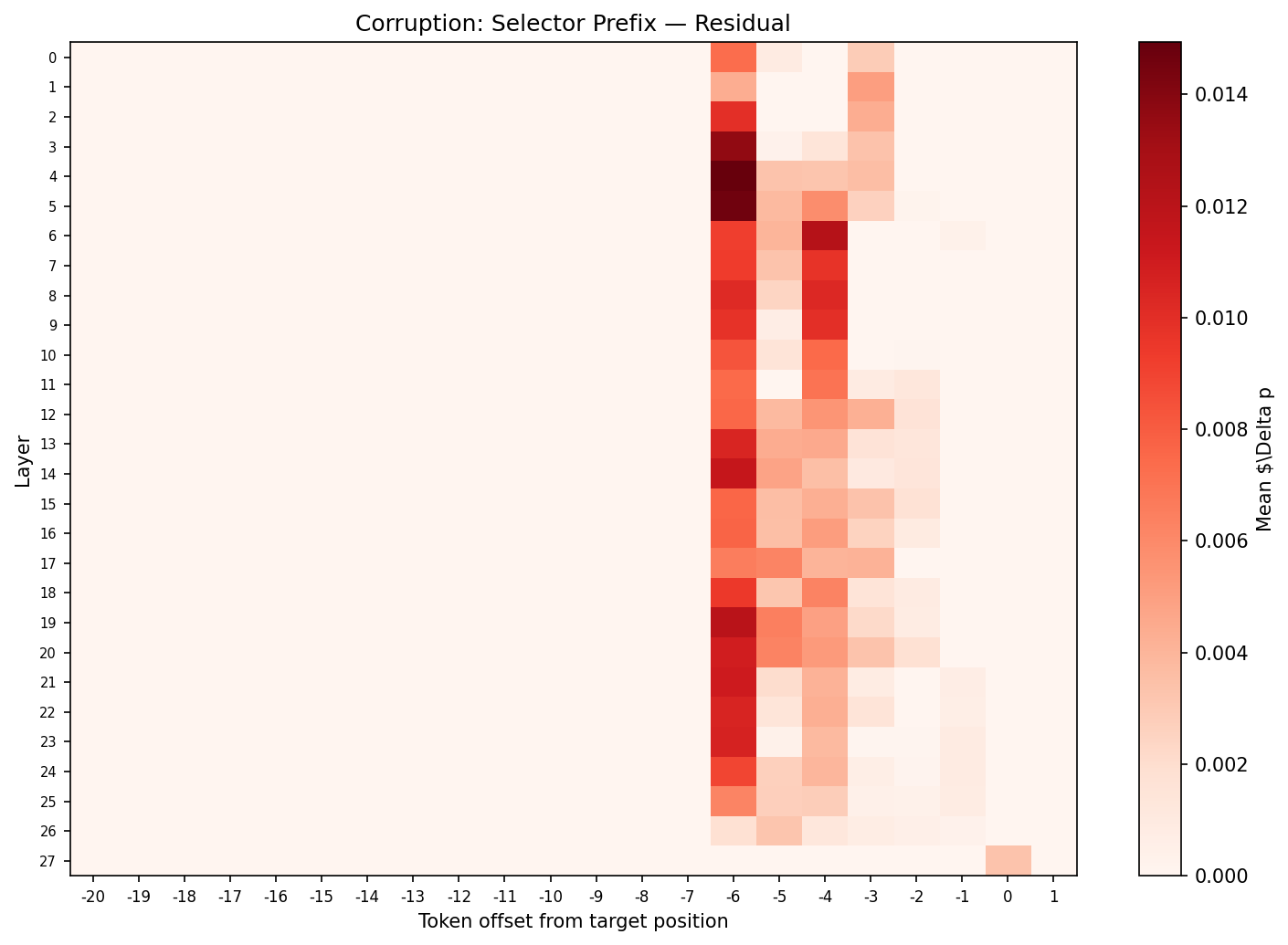}\hfill
    \includegraphics[width=0.32\linewidth]{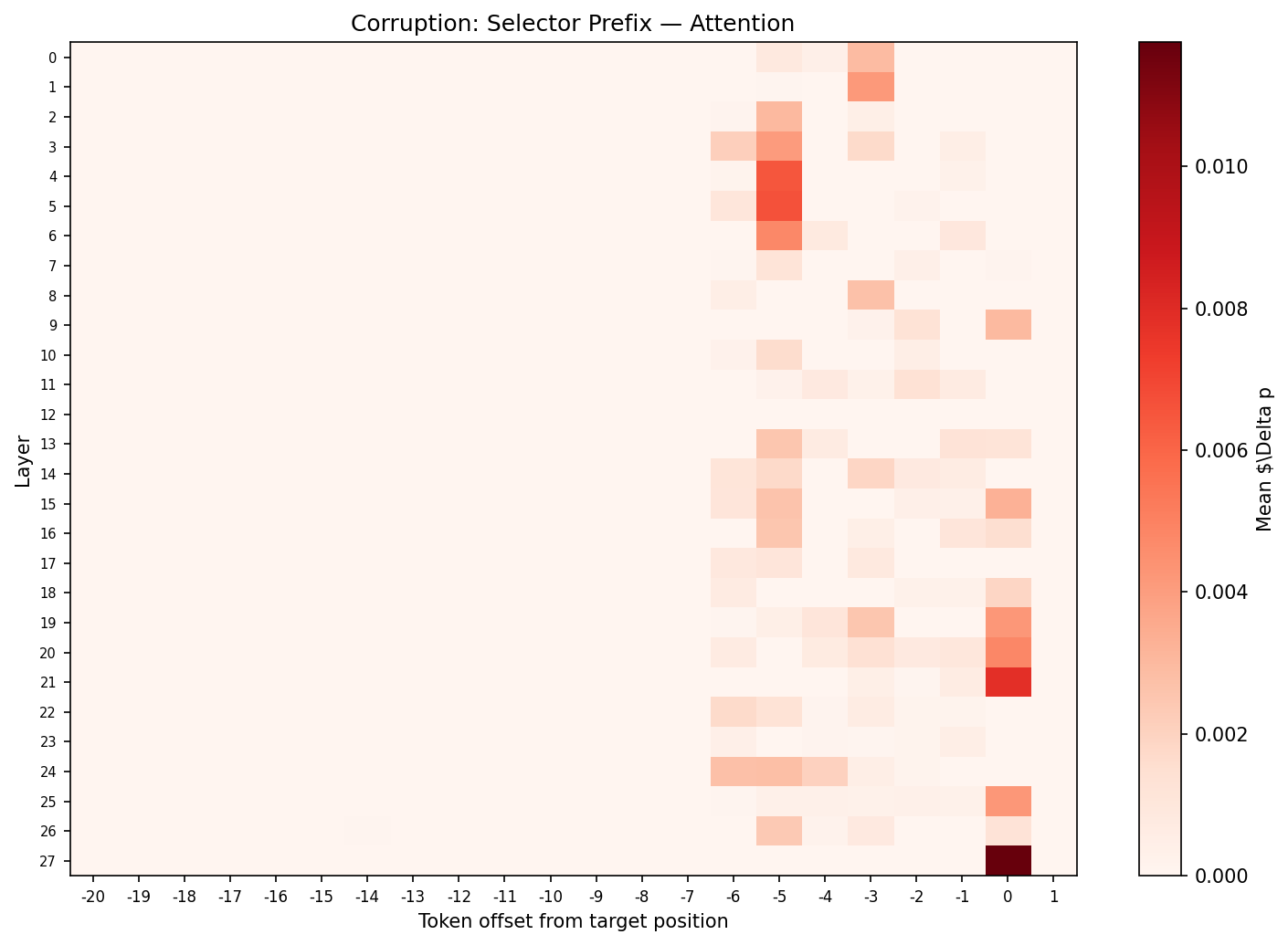}\hfill
    \includegraphics[width=0.32\linewidth]{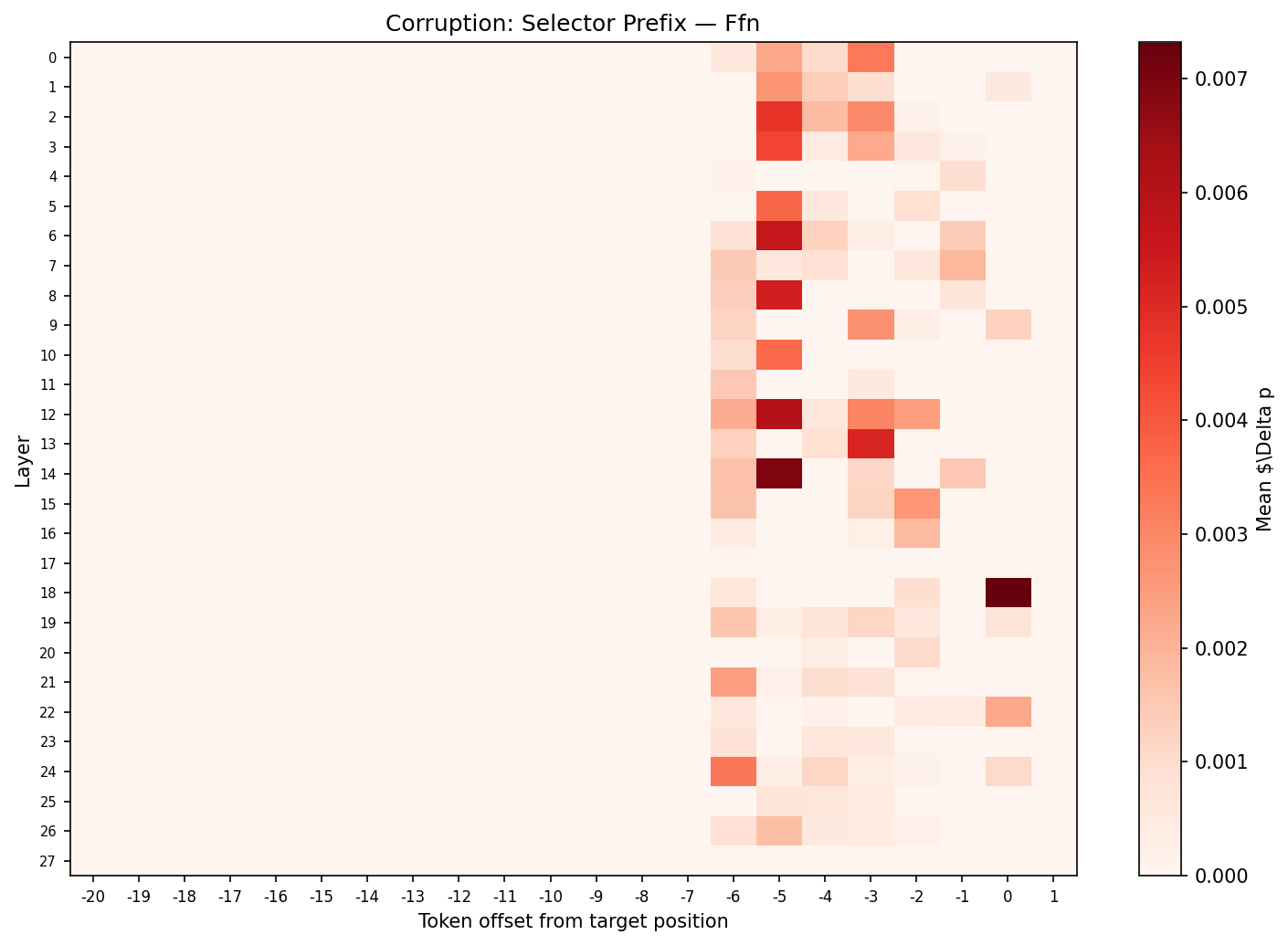}
    \caption{Full \texttt{Search} causal trace with corruption span = selector start (left: residual, middle: attention, right: FFN).}\label{fig:search-trace-1}
\end{figure*}

\begin{figure*}[t]
    \centering
    \includegraphics[width=0.32\linewidth]{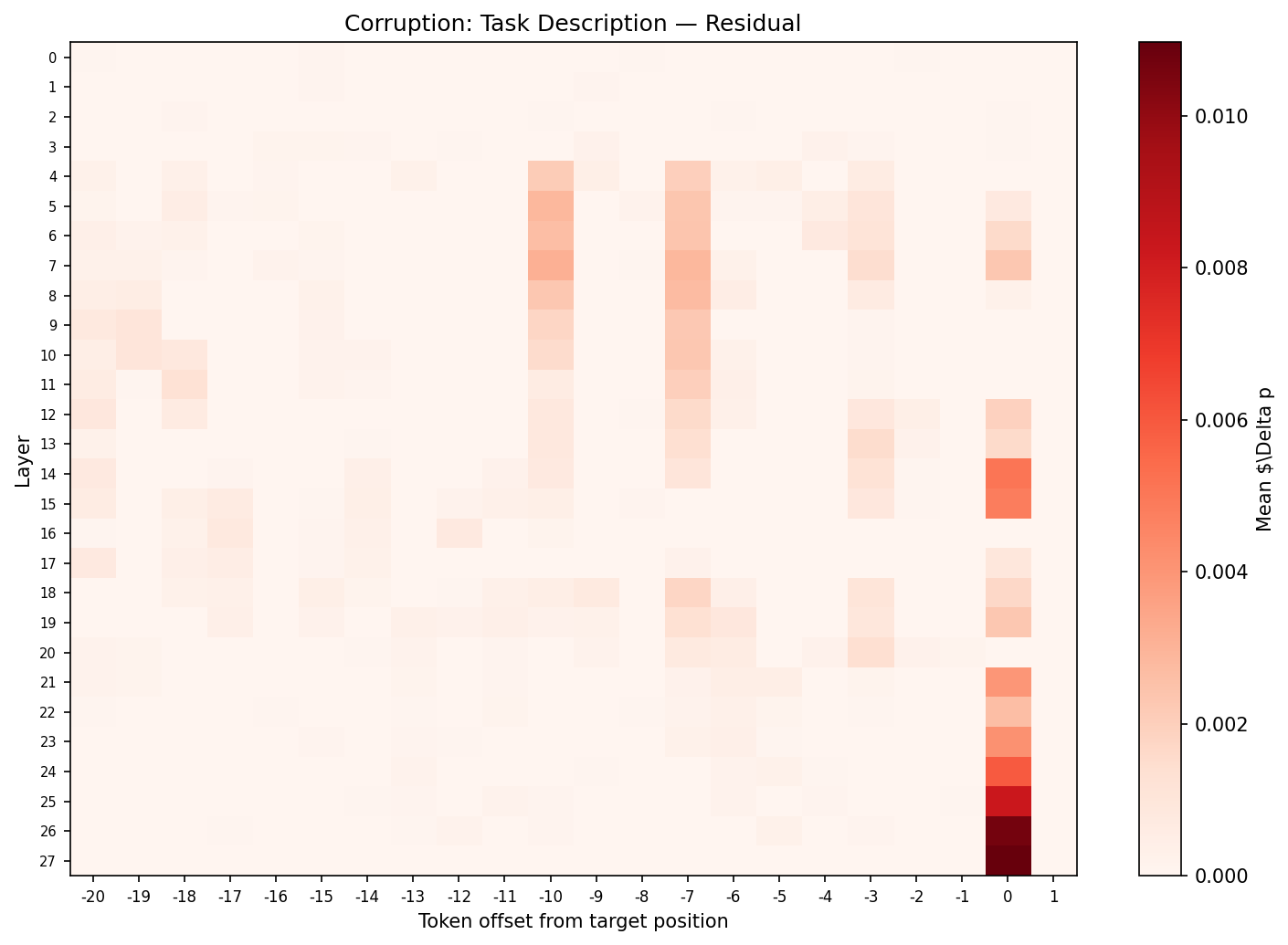}\hfill
    \includegraphics[width=0.32\linewidth]{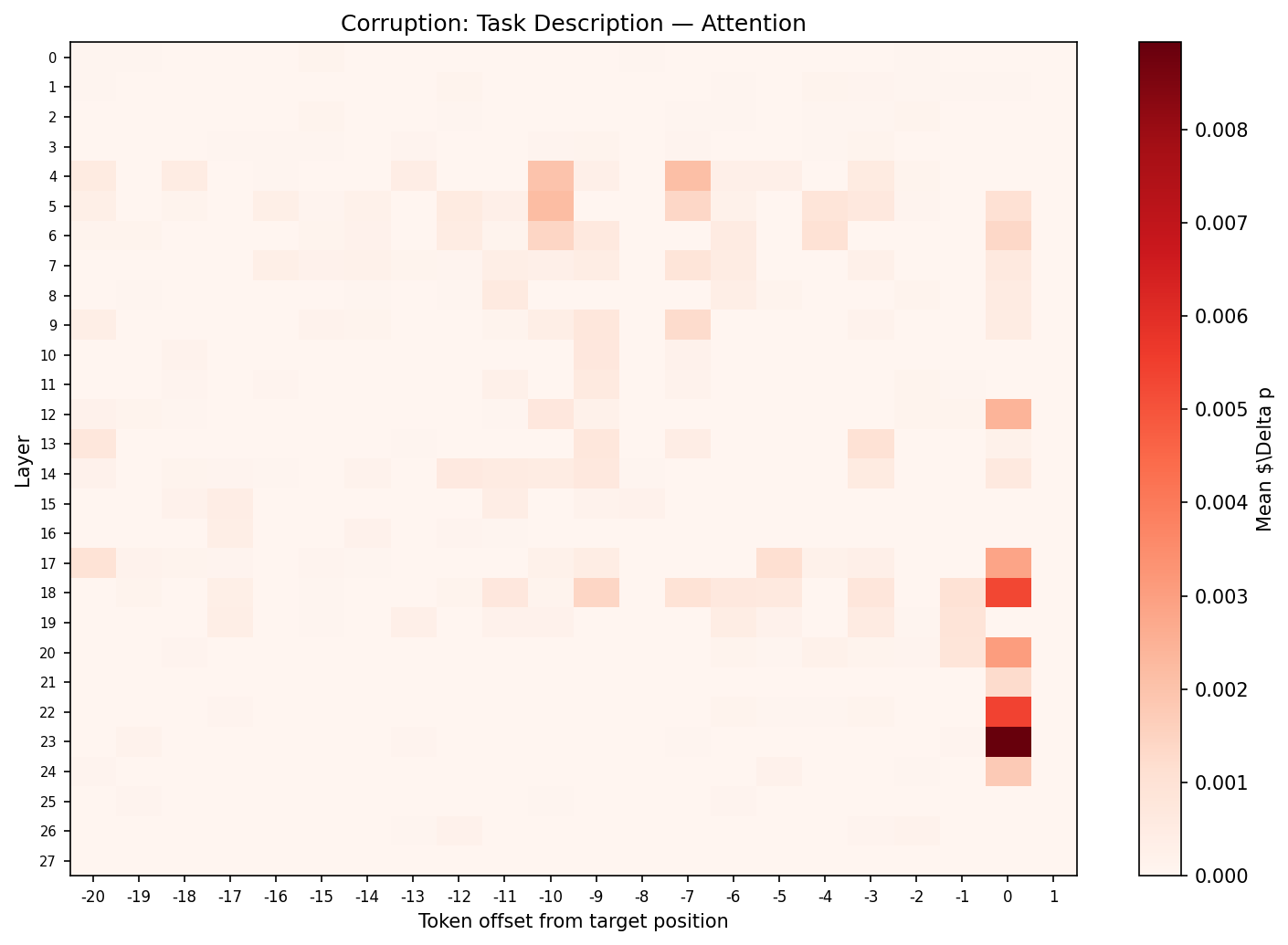}\hfill
    \includegraphics[width=0.32\linewidth]{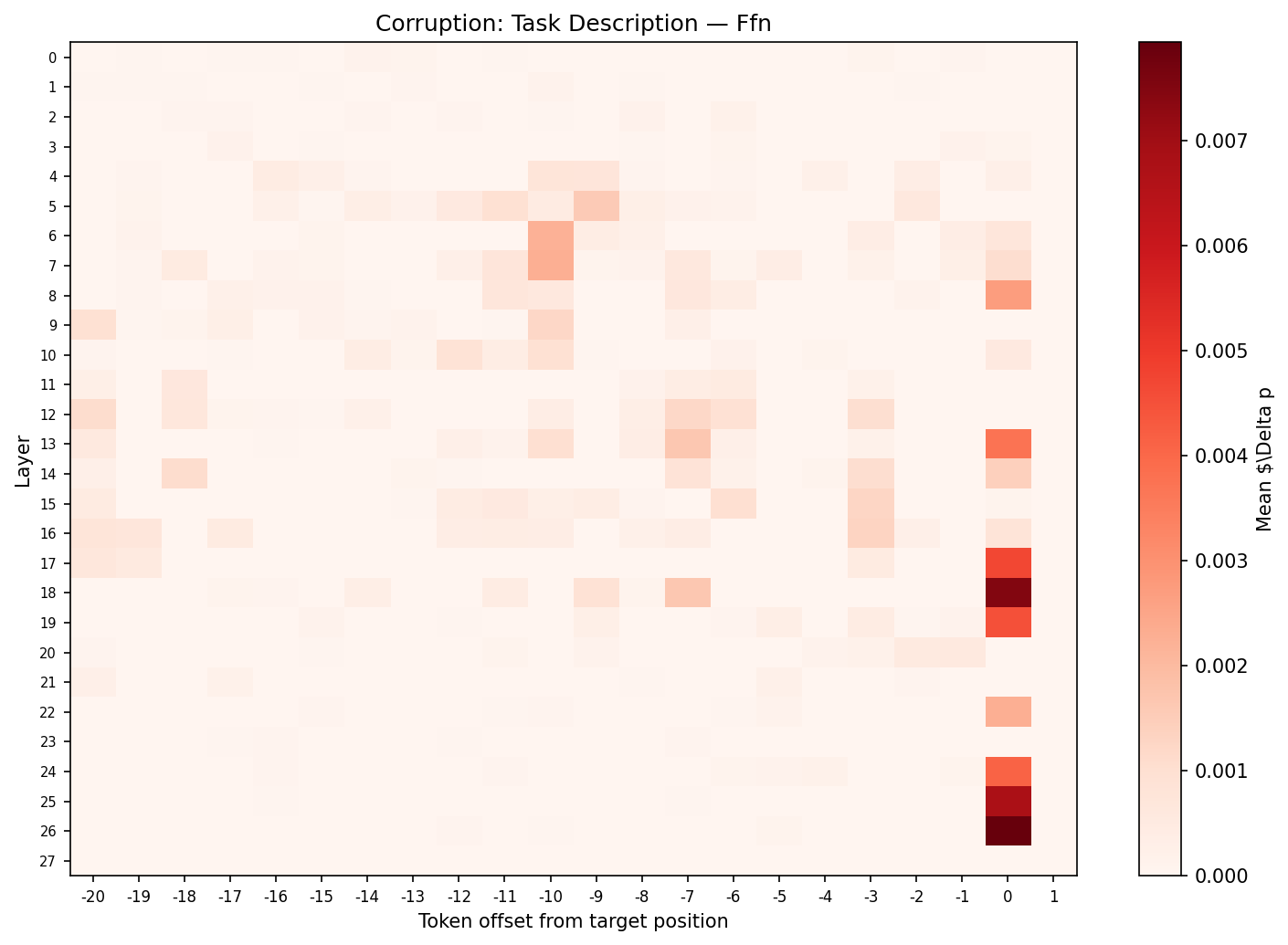}
    \caption{Full \texttt{Search} causal trace with corruption span = task description (left: residual, middle: attention, right: FFN).}\label{fig:search-trace-2}
\end{figure*}

\begin{figure*}[t]
    \centering
    \includegraphics[width=0.32\linewidth]{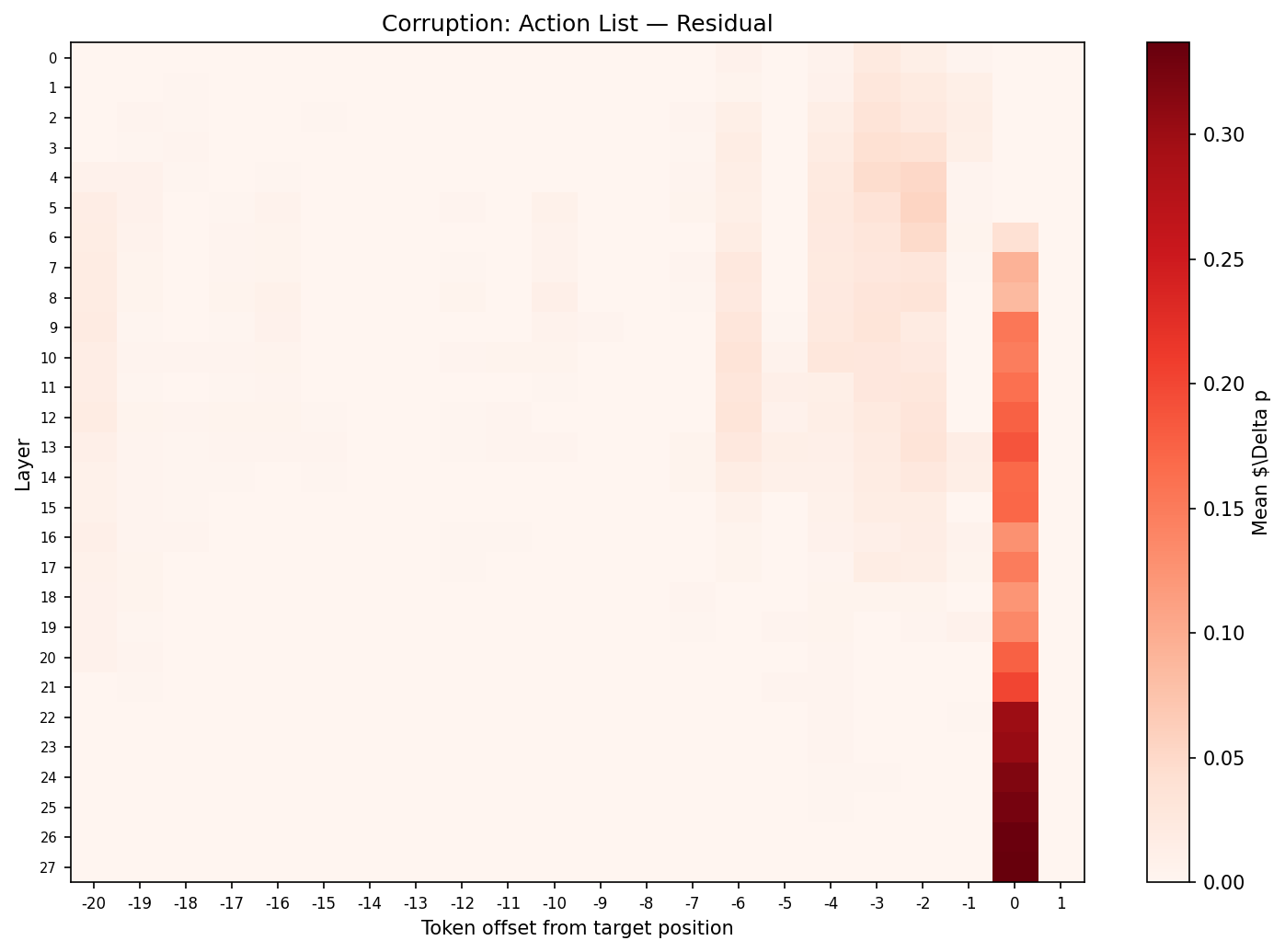}\hfill
    \includegraphics[width=0.32\linewidth]{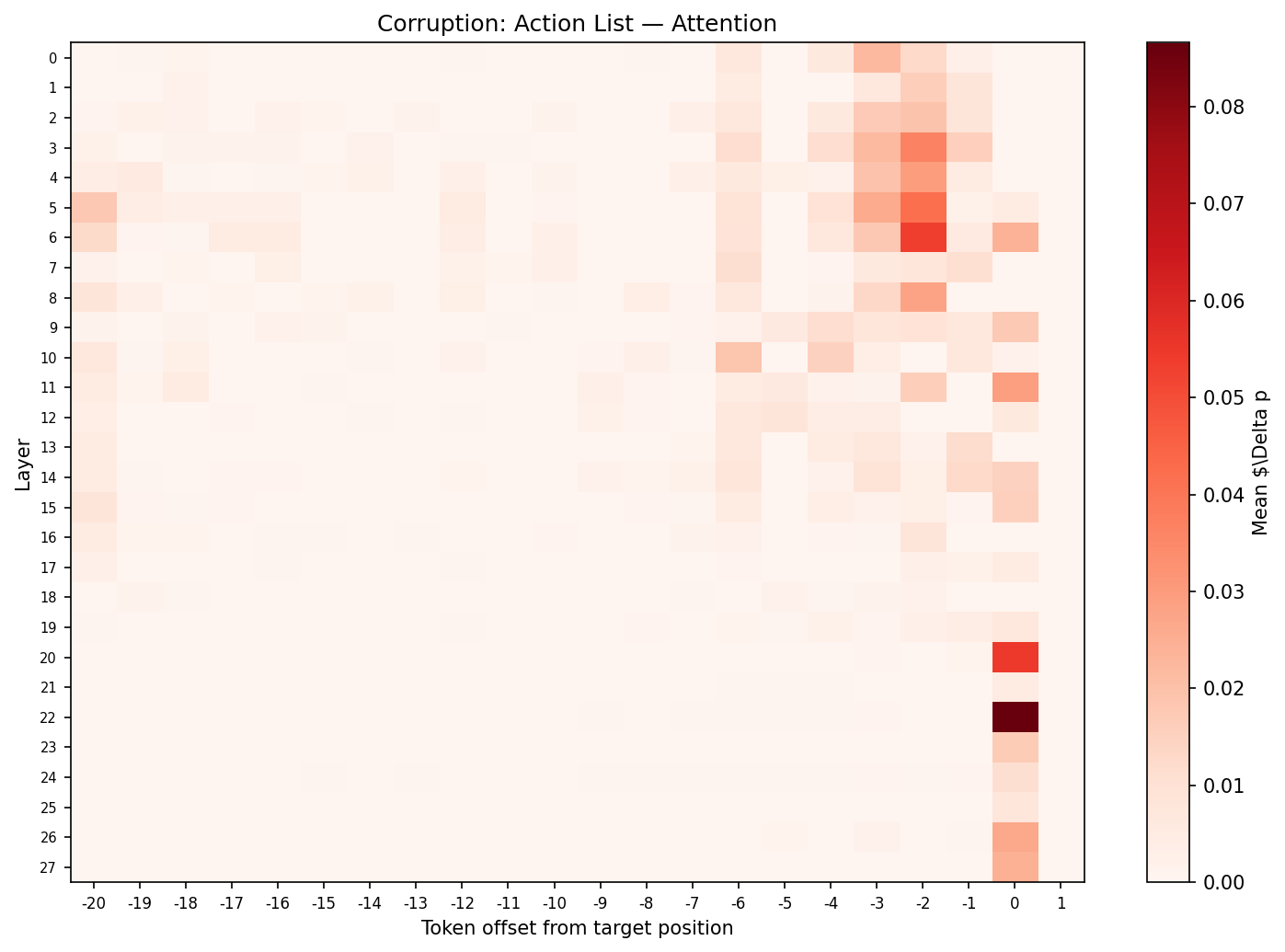}\hfill
    \includegraphics[width=0.32\linewidth]{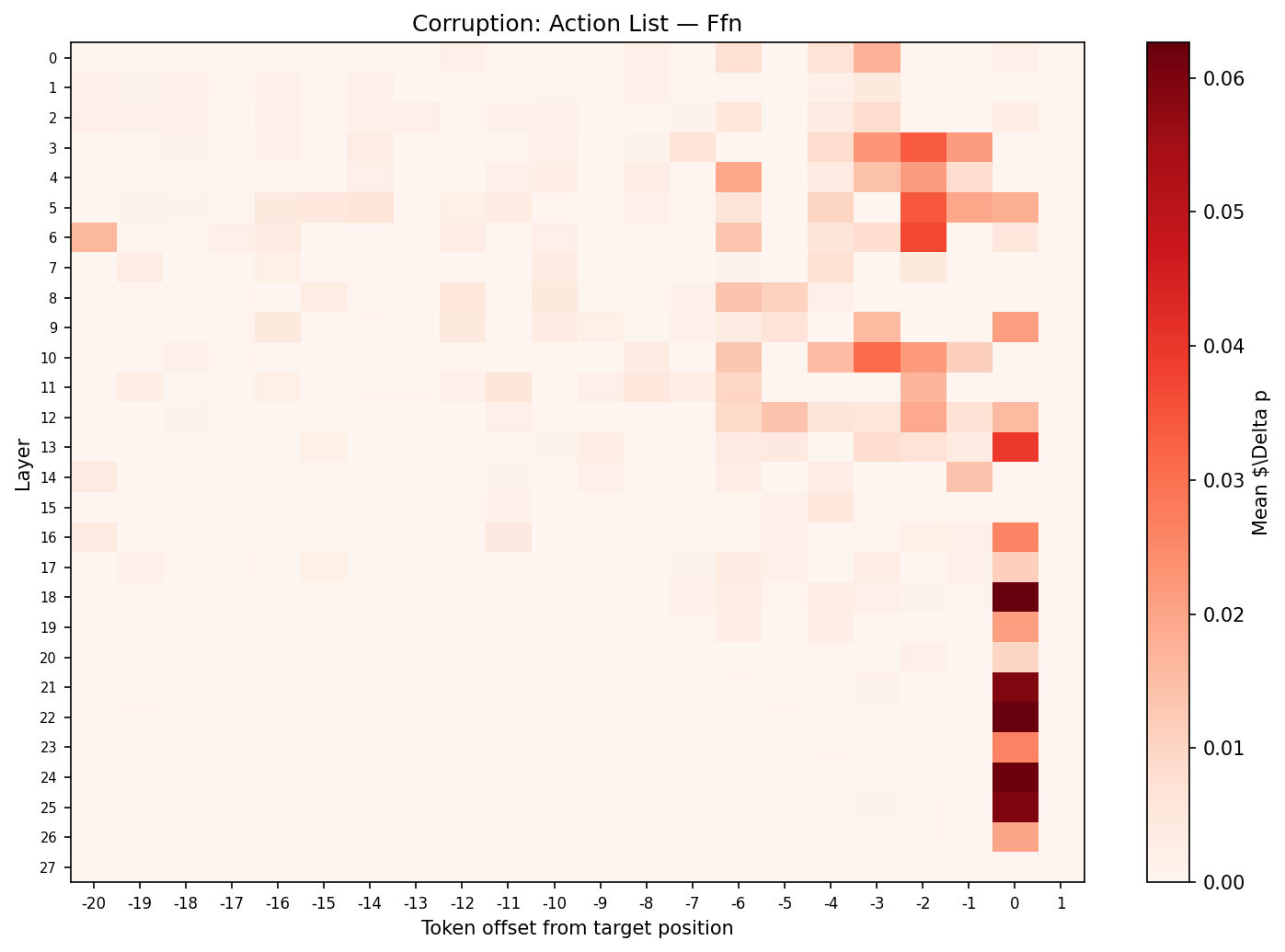}
    \caption{Full \texttt{Search} causal trace with corruption span = action list (left: residual, middle: attention, right: FFN).}\label{fig:search-trace-3}
\end{figure*}

\section{Counterfactual Trajectory Divergence}
\label{app:counterfactual_divergence}

\begin{table}[H]
\centering
\caption{Trajectory divergence after removing the chosen action at step 0.
\emph{Sel.\ match}: \% of pairs where original and CF trajectories choose the same action.
\emph{DOM Jac.}: word-level Jaccard similarity between the pages the two trajectories land on.
$\Delta$ = CF hall.\ rate $-$ orig.\ hall.\ rate.}
\resizebox{\columnwidth}{!}{%
\begin{tabular}{cccc}
\toprule
\textbf{Step} & \textbf{Sel.\ match (\%)} & \textbf{DOM Jac.} & $\boldsymbol{\Delta}$ \textbf{(\%)} \\
\midrule
0 & --   & 1.000 & $+$38.0 \\
1 & 28.1 & 0.521 & $-$8.5  \\
2 & 22.2 & 0.634 & $-$6.7  \\
3 & 27.3 & 0.672 & $+$12.1 \\
\bottomrule
\end{tabular}}
\label{tab:trajectory_divergence}
\end{table}

To illustrate the stationarity violation described in Section~\ref{sec:web-agent-resutls}, we attempt to construct counterfactuals by removing the chosen valid action from the action list and re-running the model. Table~\ref{tab:trajectory_divergence} summarises the results. Two key issues emerge. First, only 38\% of such perturbations lead to a hallucinated output at step 0, meaning that removing the chosen action does not reliably produce the contrastive behaviour needed for activation patching; hallucination cannot be isolated to the chosen action. Second, the trajectory diverges substantially in subsequent steps: selector agreement between the original and counterfactual runs drops to around 25\% by step 1 and the DOM Jaccard similarity between the pages they land on falls to 0.52, indicating that the two runs are navigating entirely different parts of the website within one step. As a result, representations built from the respective states of the two runs at step 2 or beyond cannot be meaningfully swapped, undermining the paired-trial assumption that activation patching requires.

\section{Probe-Based Representation Steering}
\label{app:steering}

To further test whether the hallucination-predictive direction in the residual stream is causal to the output, we apply representation steering. For each position and a set of candidate layers, we compute a steering vector as the normalised probe weight in un-normalised activation space:
\[
  \hat{\mathbf{v}} = -\frac{\mathbf{w}/\boldsymbol{\sigma}}{\|\mathbf{w}/\boldsymbol{\sigma}\|},
\]
where \(\mathbf{w}\) is the trained linear probe weight and \(\boldsymbol{\sigma}\)
is the per-feature standard deviation from the training split.
The sign is chosen so that \(\hat{\mathbf{v}}\) points toward the
\emph{valid} side of the probe's decision boundary. We sweep \(\alpha \in \{10, 20, 40, 60, 80\}\) and report
\textbf{fix rate} (fraction of confirmed hallucinated cases that become valid)
and \textbf{break rate} (fraction of confirmed valid cases that become hallucinated).

Table~\ref{tab:steering_main} shows results for all six (position, layer) configurations.
\texttt{selector\_start} at layers 17 and 18 are the most effective, achieving fix rates of 85.4\% and 82.9\% respectively at \(\alpha=40\)
with only 7\% break rate. \texttt{prompt\_end} and \texttt{selector\_value\_start} are substantially weaker.

\begin{table*}[t]
\centering
\caption{Fix and break rates for probe-based steering across (position, layer)
configurations. Each cell shows fix\,/\,break\,\%. Bold marks the best
fix rate for each configuration.}
\label{tab:steering_main}
\setlength{\tabcolsep}{4pt}
\begin{tabular}{llrrrrr}
\toprule
\textbf{Position} & \textbf{Layer} &
  $\boldsymbol{\alpha=10}$ & $\boldsymbol{\alpha=20}$ &
  $\boldsymbol{\alpha=40}$ & $\boldsymbol{\alpha=60}$ & $\boldsymbol{\alpha=80}$ \\
\midrule
\texttt{prompt\_end}            & 18 & 14.6\,/\,0.8  & 36.6\,/\,1.5  & \textbf{41.5}\,/\,10.8 & 29.3\,/\,29.5 & 0.0\,/\,100.0 \\
\texttt{prompt\_end}            & 26 &  0.0\,/\,0.0  &  2.4\,/\,2.3  &  \textbf{4.9}\,/\,3.1  &  4.9\,/\,3.1  & 12.2\,/\,2.3  \\
\texttt{selector\_start}        & 17 & 43.9\,/\,0.0  & 75.6\,/\,3.1  & \textbf{85.4}\,/\,7.0  & 48.8\,/\,30.2 &  0.0\,/\,96.9 \\
\texttt{selector\_start}        & 18 & 34.2\,/\,3.1  & 56.1\,/\,2.3  & 82.9\,/\,7.0           & \textbf{85.4}\,/\,14.0 & 2.4\,/\,91.5 \\
\texttt{selector\_value\_start} & 18 & 14.6\,/\,4.6  & 46.3\,/\,10.1 & \textbf{46.3}\,/\,19.4 & 12.2\,/\,57.4 &  0.0\,/\,100.0 \\
\texttt{selector\_value\_start} & 21 &  9.8\,/\,3.9  & 14.6\,/\,7.0  & 17.1\,/\,29.5          & \textbf{29.3}\,/\,33.3 & 14.6\,/\,40.3 \\
\bottomrule
\end{tabular}
\end{table*}

To further test whether the steering direction encodes a causal and untangled hallucination direction, we also evaluate across prompt formats,
we apply the best-performing vector (\texttt{selector\_start} L17, trained on the original JSON action-list format) to the OOD numbered-list prompt
format described in Appendix~\ref{app:ood-prompt}, using the same alpha sweep. Table~\ref{tab:steering_ood} compares the two formats side by side.

\begin{table}[H]
\centering
\caption{Fix and break rates for \texttt{selector\_start} L17 steering on
the original JSON format vs.\ the OOD numbered-list format. The same
steering vector is used in both conditions.}
\label{tab:steering_ood}
\setlength{\tabcolsep}{4pt}
\begin{tabular}{lrrr}
\toprule
& $\boldsymbol{\alpha=10}$ & $\boldsymbol{\alpha=20}$
& $\boldsymbol{\alpha=40}$ \\
\midrule
Original format & 43.9\,/\,0.0 & 75.6\,/\,3.1 & \textbf{85.4}\,/\,7.0 \\
OOD format      & 19.4\,/\,0.0 & 41.9\,/\,2.3 & \textbf{58.1}\,/\,7.0 \\
\midrule
$\Delta$ fix    & $-$24.5 & $-$33.7 & $-$27.3 \\
\bottomrule
\end{tabular}
\end{table}

Fix rate drops by $\sim$27 percentage points at \(\alpha=40\), with break rate nearly identical across formats.
The consistent gap across all alpha values indicates that the steering effect is entangled with prompt format rather than with the
hallucination direction by itself, which is also the result observed in the probe generalisation results (Figure~\ref{fig:ood-probes}).

\section{Out-of-Distribution Prompt Format}
\label{app:ood-prompt}

The following prompts are used for the distributional mismatch evaluation (Figure~\ref{fig:ood-probes}): we run the model on the same 300 tasks with identical selectors but this different surrounding syntax, and collect activations at the same three positions. The available action list is reformatted from a JSON array into a numbered markdown list; the model is instructed to respond in the same list format rather than JSON.

\paragraph{System message.}
\begin{tcblisting}{prompttcb}
You are a web navigation agent. Always respond with exactly two lines: [THOUGHT] and [ACTION]. The [ACTION] must use the numbered list format: [ACTION] [TYPE] "visible text"
   selector: <selector exactly as shown>
\end{tcblisting}

\paragraph{User message.}
Variables \texttt{<TASK>}, \texttt{<DOM>}, and \texttt{<ACTIONS\_LIST>} denote the dataset task string, the DOM summary, and the numbered action list, respectively.
\begin{tcblisting}{prompttcb}
Task: <TASK>

You are given a task to perform on a webpage. Propose the next step that is helpful towards achieving the task.

You should output only one action. Your response must follow this exact format:

[THOUGHT] <one short sentence>
[ACTION] [ACTION_TYPE] "visible text on element"
   selector: <selector exactly as shown in the list>
   value: <only include this line for TYPE or SELECT actions>

Below is the simplified representation of the current state of the webpage:
<DOM>

Below is the list of available actions. DO NOT output an action not in this list. Copy the selector EXACTLY as shown, character for character.

Available Actions:
1. [CLICK] "visible text"
   selector: <xpath>

2. [TYPE] "visible text"
   selector: <xpath>
   value: <value>

...
=========END OF ACTIONS=========

Continue with:
[THOUGHT] <one short sentence>
[ACTION] [ACTION_TYPE] "visible text"
   selector: <selector exactly as shown>
\end{tcblisting}

\section{SAE Analysis}\label{app:sae}

We apply a pretrained SAE for \texttt{Qwen2.5-7B-Instruct}\footnote{\url{https://huggingface.co/andyrdt/saes-qwen2.5-7b-instruct}} using a BatchTopK architecture with a dictionary of 131{,}072 features and a sparsity of $k{=}64$ active features per token. We analyze layer 27 (the layer with the highest causal recovery in Table~\ref{tab:causal-tracing-search}) at the \texttt{selector\_value\_start} position. For each instance, we pass the residual stream representation through the SAE and take the maximum activation across all tokens per feature, yielding a scalar indicating whether each feature activates at least once at that position. We then average these values across instances within each class (hallucinated vs. valid) and identify which tokens each feature most promotes via its dot product with the unembedding matrix. As shown in Figure~\ref{fig:sae}, the top-10 features by absolute difference between classes mostly promote tokens unrelated to the observed hallucination behavior and do not yield an interpretable explanation. This highlights how SAEs pretrained on text can struggle to encode features that emerge in agent interaction.

\section{Attributions and Licenses}

\paragraph{Model.}
All experiments use \texttt{Qwen2.5-7B-Instruct} \citep{qwen2025qwen25technicalreport}, released under the Apache 2.0 license.

\paragraph{Dataset.}
Agent trajectories are collected on Online-Mind2Web \citep{xue2025an}, released under a Creative Commons Attribution 4.0 International (CC-BY~4.0) license. Use of this dataset also requires attribution to the original Mind2Web benchmark \citep{deng2023mind2web}. We use this dataset in accordance with its intended use, which consists of web agent evaluation for research purposes.

\paragraph{Sparse Autoencoder.}
We use pretrained SAE checkpoints for \texttt{Qwen2.5-7B-Instruct} from \url{https://huggingface.co/andyrdt/saes-qwen2.5-7b-instruct}, released under the Apache 2.0 license.

\paragraph{Analysis Infrastructure.}
We use TransformerLens \citep{nandatransformerlens2022}, released under the MIT license for activation caching and hook-based interventions.  We also use common open-source scientific libraries, namely PyTorch (BSD-3-Clause), Hugging Face Transformers (Apache 2.0), scikit-learn (BSD-3-Clause), NumPy (BSD-3-Clause), and Matplotlib (PSF).

\paragraph{Compute Resources.} All experiments and trajectory collection were run on a single NVIDIA A100 80GB GPU.

\end{document}